\documentclass[runningheads]{llncs}
\usepackage{graphicx}

\usepackage{comment}
\usepackage{amsmath,amssymb} 
\usepackage{color}

\usepackage[accsupp]{axessibility}  
\usepackage{ctable} 
\usepackage{mathrsfs}
\usepackage[symbol*]{footmisc}
\usepackage{multirow}
\usepackage{tablefootnote}
\usepackage{float}
\usepackage{color}
\usepackage{subcaption}
\definecolor{commentcolor}{RGB}{62,128,128}   
\definecolor{pyin}{RGB}{216,46,128}
\newcommand{\PyComment}[1]{\ttfamily\textcolor{commentcolor}{\# #1}}  
\newcommand{\PyCode}[1]{\ttfamily\textcolor{black}{#1}} 
\newcommand{\Py}[1]{\ttfamily\textcolor{pyin}{#1}}
\usepackage{epsfig}
\usepackage{graphicx}
\usepackage{algorithm}

\begin{document}
\pagestyle{headings}
\mainmatter
\def\ECCVSubNumber{100}  

\title{Rare Wildlife Recognition with Self-Supervised Representation Learning} 

\titlerunning{Master Thesis, ETH Zürich, Switzerland}
%
\author{Master Thesis}
\authorrunning{X. Zheng.}
%
\institute{Xiaochen Zheng\\
\email{xzheng@student.ethz.ch} \\ ETH Zürich, Switzerland}
\maketitle

\begin{abstract}
Automated animal censuses with aerial imagery are a vital ingredient towards wildlife conservation. Recent models are generally based on supervised learning and thus require vast amounts of training data. Due to their scarcity and minuscule size, annotating animals in aerial imagery is a highly tedious process. In this project, we present a methodology to reduce the amount of required training data by resorting to self-supervised pretraining. In detail, we examine a combination of recent contrastive learning methodologies like Momentum Contrast (MoCo) and Cross-Level Instance-Group Discrimination (CLD) to condition our model on the aerial images without the requirement for labels. We show that a combination of MoCo, CLD, and geometric augmentations outperforms conventional models pretrained on ImageNet by a large margin. Meanwhile, strategies for smoothing label or prediction distribution in supervised learning have been proven useful in preventing the model from overfitting. We combine the self-supervised contrastive models with image mixup strategies and find that it is useful for learning more robust visual representations. Crucially, our methods still yield favorable results even if we reduce the number of training animals to just 10\%, at which point our best model scores double the recall of the baseline at similar precision. This effectively allows reducing the number of required annotations to a fraction while still being able to train high-accuracy models in such highly challenging settings.
\supervision{Dr. Benjamin Kellenberger, ECEO, EPFL, Switzerland\\ Prof. Dr. Devis Tuia, ECEO, EPFL, Switzerland\\Prof. Dr. Irena Hajnsek, EO, ETH Zürich, Switzerland}
\submission{August 2021\footnote{\url{https://eo.ifu.ethz.ch/studium/master-thesis.html}}}
\end{abstract}

\clearpage

\section{Introduction}
\label{intro}
\textbf{Background} Biodiversity and ecosystem services are of fundamental importance to the earth and to human society. However, human activities are causing rapid habitat destruction and environmental degradation, making biodiversity disappear at an unprecedented rate, which has reached over 1,000 times the rate of natural disappearance of species~\cite{balmford2005convention}. This requires adaptive technologies for effective environmental monitoring, which includes wildlife censuses~\cite{linchant2015unmanned}.

Wildlife censuses plays an increasingly significant role in protecting endangered species. The censuses help determine the exact number and the spatial-temporal distribution of wild animals, which is vital to assess living conditions and potential survival risks of wildlife species~\cite{Beni_2018_RSE,balmford2005convention,biodiversity_review}. Meanwhile wildlife censuses help monitor usual changes in wildlife communities~\cite{linchant2015unmanned}, which are essential to ecological balance and biodiversity. 

Manual surveys are traditional approaches of wildlife censuses in remote areas. However, serious deficiencies, such as health and life risk on human operators, low accuracies and small area coverage, are caused by using manned aircrafts, camera traps, or road driving for field work. One of the effective and feasible ways to address these issues is to replace manual surveys of wildlife reserves with counts derived automatically from images acquired by Unmanned Aerial Vehicles (UAVs), paired with deep learning models to automate the wildlife recognition task~\cite{Beni_2018_RSE,Beni_al,Devis_detecting,Kellenberger_2019_CVPR_Workshops,eikelboom2019improving}.

Supervised deep learning models have been shown to perform exceptionally well for natural images. These models are typically pretrained on large-scale curated datasets such as ImageNet~\cite{imagenet_cvpr09}, MS-COCO~\cite{coco} with supervision and then fine-tuned on the target natural imagery. Irrespective of the type of fine-tuning, this last step requires thousands of animals to be annotated, and hence expert knowledge. Meanwhile, large UAV campaigns can generate images in high numbers~\cite{Beni_2018_RSE}, with many containing either large numbers of animals, or none~\cite{Kellenberger_2019_CVPR_Workshops}, both of which cases are cumbersome for manual annotators. Furthermore, the vastness of wildlife reserves mean that wildlife is a rare sight and background dominates the majority of images. This causes two problems: (1) the datasets are strongly imbalanced toward the absence class; (2) the objects of interest are extremely small compared with the original aerial image size, illustrated in Figure. These two issues significantly downgrade the capacity of a supervised model to solve the recognition task~\cite{Beni_2018_RSE}. Hence, methods to reduce the labelling requirement are urgently needed.

\begin{figure}[t]
\begin{center}
  \includegraphics[width=0.48\linewidth]{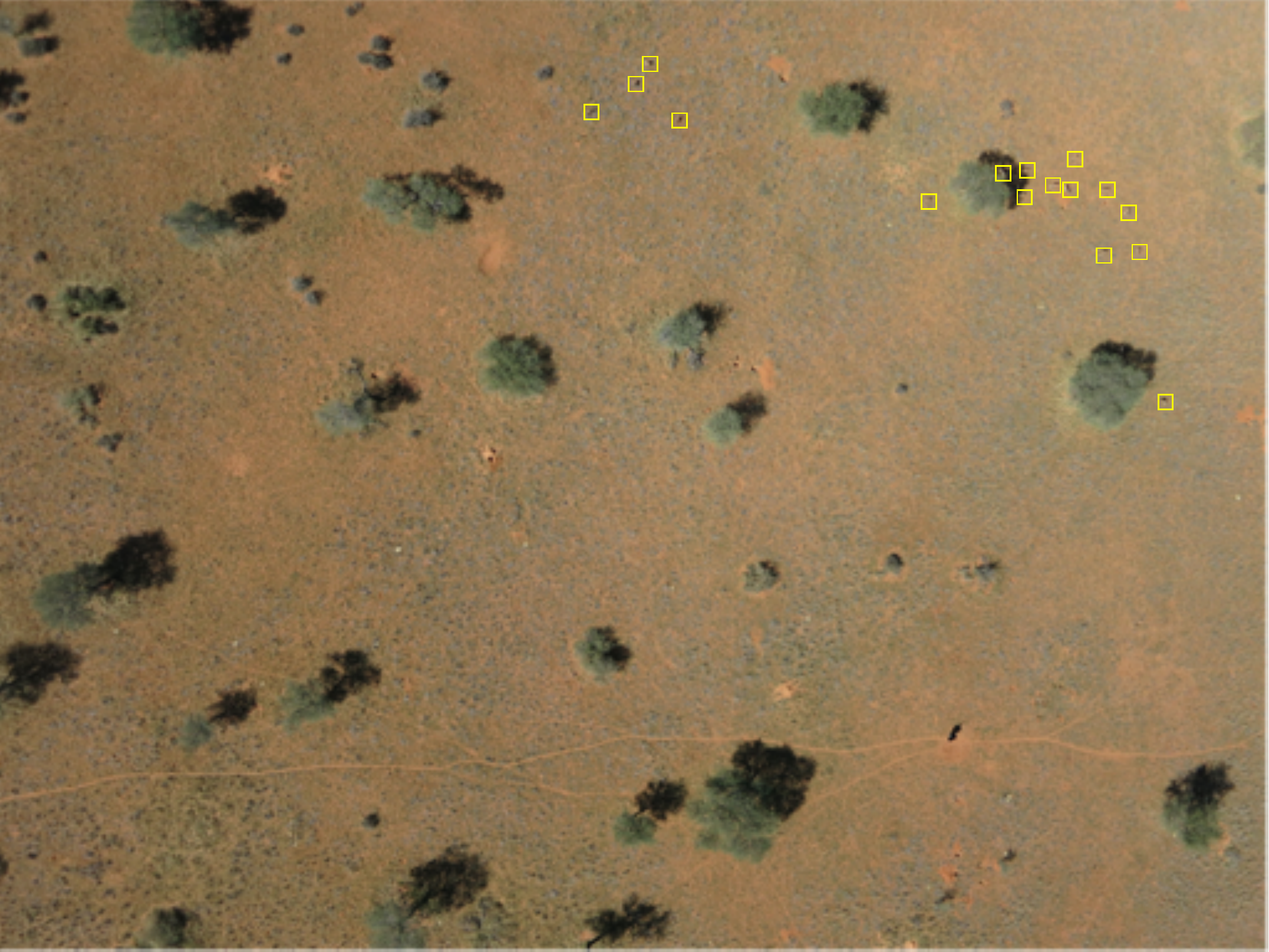}
\end{center}
  \caption{Wildlife in raw image is tiny and is a rare sight.}
\label{fig:1mfs}
\end{figure}

A promising direction to this end is to use self-supervised learning (SSL), where models are first trained in a \emph{pretext} task on the target images without the need for manually provided labels, and then fine-tuned on the actual objective (\emph{downstream task}) with manual labels~\cite{invariant}. Earlier \emph{pretext} tasks required models to reconstruct transformations between different \emph{views} of the same image. Recently, focus has shifted to contrastive learning. Here, augmentation as a method of image transformation is still employed, but with a different objective: while traditional SSL methodologies forced the model to learn representations within one data point to \emph{e.g.}, in-paint cut-out regions~\cite{inpainting,zhang2016colorful}, contrastive learning employs transformations in a comparison scheme and encourages the model to learn representations by maximizing the similarity between two randomly augmented \emph{views} of the same data point, resp. image (positive pairs) and dissimilarity between different data points (negative pairs)~\cite{Dahua_2018_CVPR,he2020momentum,mocov2,cmc,SimCLR,byol,PIRL_2020_CVPR}. The different \emph{views} of same instance are randomly generated from a stochastic data augmentation module. Recent works have identified a stronger augmentation strategy to be vital for improved learning~\cite{mocov2}. However, augmentation functions need to be carefully selected with respect to the problem and data at hand~\cite{LooC}. Choosing inadequate functions may result in removal of important information (\emph{e.g.}, random resized cropping may remove animals at the border of UAV images). In contrast, some functions benefit certain scenarios more than others (\emph{e.g.}, random vertical flips and rotations may be of limited use with natural images, but may provide strong learning signals for view-independent aerial images).

However, applying self-supervised learning techniques, such as contrastive learning, to UAV images on wildlife is challenging. One such problem is the requirement of contrastive learning methods to receive dissimilar imagery. In UAV acquisitions, as illustrated in Figure~\ref{fig:correlation}, high image sampling frequencies will generate strong autocorrelations between acquisitions in short time intervals (similar to adjacent frames in a video). Also, the vastness of many wildlife areas result in consistent characteristics, and hence in repeated or similar patches in the dataset. Many instance discrimination based methods, such as NPID~\cite{Dahua_2018_CVPR}, MoCo~\cite{he2020momentum,mocov2}, and SimCLR~\cite{SimCLR}, are based on the assumption that each instance is significantly different from others and that each instance can be treated as a separate category. The large similarity between training images mean that the negative pairs used in the contrastive learning process is likely to be composed of highly similar instances, which in turn compromises feature representations due to incorrect repulsion between similar images. Instead of exploring the effect of hard negative sampling~\cite{hard_neg_mix,hard_neg_sample}, in this paper we propose to solve these problems with Cross-Level Instance-Group Discrimination (CLD)~\cite{Wang_2021_CVPR}, which aims to deal with highly correlated datasets. Furthermore, we hypothesize that top view UAV imagery should be \emph{invariant} to geometric transformations, \emph{e.g.}, rotation. From this perspective, we propose to apply extra geometric transformation to contrastive model, which captures invariant information of UAV images introduced by different augmentations. 

\textbf{Scope and Objective.} The thesis is mainly focus on four aspects:

(1) Processing and analyzing the raw data from UAVs. The raw images are large in size and have very high resolution. Due to the limited computation capacity, the deep models can only handle with patches in appropriate sizes cropped from raw images. The processing aims to find a cropping method without using annotation information that adapts to self-supervised pretraining.

(2) Defining the downstream task. We aim to exploit how SSL pretraining will perform on object recognition task. Thus for downstream task, all patches should be image-level annotated and the cropping methods should be different from the one of SSL pretraining.

(3) Designing domain-specific algorithm. We aims to design algorithms based on the characteristics of our natural data and to select the best fitted algorithm by applying certain criteria, \emph{e.g.}, linear protocol and classification accuracy. The features of remote sensing images in latent space should be rotation invariant. Hence, our model should learn rotation-invariant features to obtain good performance

(4) In-depth analysis of performance on downstream task. 

The quantification of the training uncertainty associated with the random seed selection is not necessary as part of the work. The study of model generalization to other datasets as well as the influence of spatial and temporal variability of data acquisition is not included in the project. The numerical computer vision algorithms are not included.

\textbf{Contributions of This Thesis.} We propose an SSL model to pretrain on our wildlife dataset by contrastive learning. We apply the pretraining process to downstream wildlife recognition task. Our work build on the work of MoCo~\cite{he2020momentum,mocov2} and Cross-Level Instance-Group Discrimination (CLD)~\cite{Wang_2021_CVPR}. The contributions of this thesis can be summarized as follows:

(1) We propose a methodology for image-level wildlife recognition with a reduced number of annotations by self-supervised pretraining.

(2) We show that using self-supervised pretraining outperforms supervised ImageNet pretraining on downstream recognition task.

(3) We design a domain-specific geometric augmentation and apply it to state-of-the-art models, \emph{e.g.}, MoCo v2~\cite{mocov2} and MoCo+CLD~\cite{Wang_2021_CVPR}.

(4) We apply image mixup strategies to self-supervised pretraining (MoCo), which are initially used in supervised learning models. We further find that MoCo with image mixup strategies outperforms ImageNet pretraining fine-tuned with all available training labels. Namely, we improve the preformance of MoCo v2 on our dataset without increasing the model parameters. Our self-supervised pretraining learns representations of natural wildlife scenes more efficiently than supervised pretraining.


\section{Related Work}

\begin{figure}[t]
\begin{center}
  \includegraphics[width=0.6\linewidth]{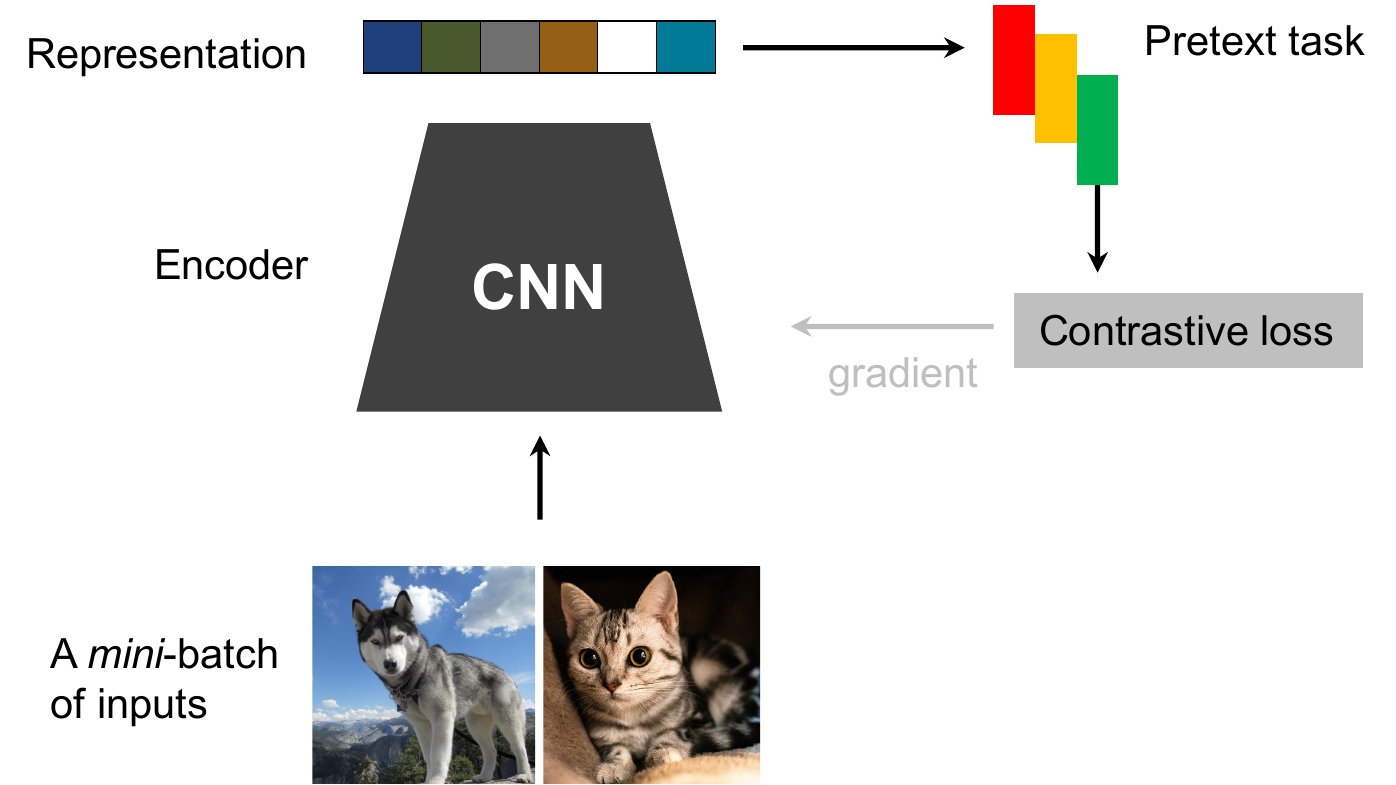}
\end{center}
  \caption{A basic framework of self-supervised representation learning. It consists of data loader, encoder, pretext task.}
\label{fig:ssl}
\end{figure}

\noindent\textbf{Self-Supervised Representation Learning.} Self-Supervised representation learning in computer vision aims to learn effective visual representations of images without any supervision. Namely, Self-Supervised deep models learn high-level representations from raw image inputs. \cite{jietang_review} summarizes the self-supervised mechanisms into three categories: generative model, contrastive model, and adversarial model. In this thesis, we only focus on the contrastive model. The basic framework is illustrated in Figure~\ref{fig:ssl}.

Unlike in the field of classic computer vision, self-supervised learning of aerial images has not yet been fully studied. \cite{RMSSL_Stojnic_2021_CVPR} apply Contrastive Multiview Coding~\cite{tian2020view} to learn aerial image representations on both RGB and multispectral remote sensing images. \cite{kang2020deep} proposes a method based on contrastive learning with different image augmentations. \cite{tao2020remote} analyze different pretext tasks, \emph{e.g.}, inpainting~\cite{inpainting}, context prediction~\cite{context_prediction}, and contrastive learning with different image augmentations on remote sensing dataset. Besides, \cite{ayush2020geography} use geo-location classification as the pretext task. The encoder is trained by predicting the global geo-location of input image. \cite{land-cover} generate pseudo-labels of unlabeled UAV imagry data to improve the downstream classification accuracy. Most of the downstream tasks in aerial imaery domain are scene classification, \emph{e.g.}, land-cover or land-use~\cite{jean2019tile2vec} classification. \cite{ayush2020geography}  apply Self-Supervised learning to transferred downstream tasks, \emph{e.g.}, object detection and image segmentation. Different from those tasks, our task is more domain specific, which is tiny and rare wildlife recognition in the wild.

\noindent\textbf{Pretext tasks.} Self-Supervised representation learning is designed to solve certain pretext tasks. \cite{rotation} uses the rotation angle as pseudo label and learn underlined structure of the objects by predicting rotation angle. \cite{context_prediction,invariant,ding2021unsupervised} perform region-level relative location prediction. Other missions like color in-painting~\cite{inpainting,zhang2016colorful}, and solving jigsaw puzzles~\cite{jigsaw} are also applied as pretext tasks. In contrastive learning, learning between different augmented views are used as pretext tasks.  In this work, we apply a instance discrimination task~\cite{Dahua_2018_CVPR} in addition to geometric invariant mapping to contrastive self-supervised models.

\noindent\textbf{Contrastive Representation Learning.} Recently, the most competitive representation learning method without labels is Self-Supervised contrastive learning. Contrastive methods~\cite{SimCLR,he2020momentum,mocov2,byol,SwAV,cpc,invariant,lecun_invariant_mapping,PIRL_2020_CVPR} train a visual representation encoder by attracting positive pairs from the same instance in latent space while repulsing negative pairs from different instances. One of the most important parts in contrastive learning is the selection of positive and negative pairs~\cite{tian2020view} for instance discrimination. To create positive pairs without prior label information, one common way is to generate multiple views of input images. \cite{mocov2,he2020momentum,SimCLR} apply a stochastic augmentation module to randomly augment an input image twice. \cite{LooC} proposes a new model with multi-augmentation and multi-head. It constructs multiple embeddings and captures varying and invariant information introduced by different augmentations. Methods combined contrastive learning with online clustering~\cite{SwAV,Wang_2021_CVPR,li2021prototypical} are proposed to boost the performance of self-supervised learning which explore the data manifold to learn image representations by capturing invariant information.

\noindent\textbf{Image Mixture Strategies.} Image mixture is well defined in supervised learning. Image mixture mixes pixel values of two images. Images in mixup~\cite{zhang2018mixup} are generated by mixing pixel values of two randomly selected images. The proportion of pixel strength during data mixing is decided by the mixing factor $\lambda \in [0, 1]$. The values for $\lambda$ are drawn from the beta distribution. 
Cutmix~\cite{cutmix_2019_ICCV} is the way to generate a new training sample by locally combining two samples of their region and whole image. Different from mixup~\cite{zhang2018mixup}, this operation will replace pixels within particular locations of a region. Augmix~\cite{hendrycks*2020augmix} performs data mixing using the input image itself. It mixes itself with different views of its original image. The views are generated from different transforms, which are a series of randomly selected augmentation operations. Those operations are applied with three parallel augmentation chains. The output of these three chains are three mixed images. This new image is further mixed with the original image to generate final augmented output image. \cite{shen2020mix} explore applying both mixup and cutmix to contrastive learning framework.

\begin{figure}[t]
\begin{center}
  \includegraphics[width=\linewidth]{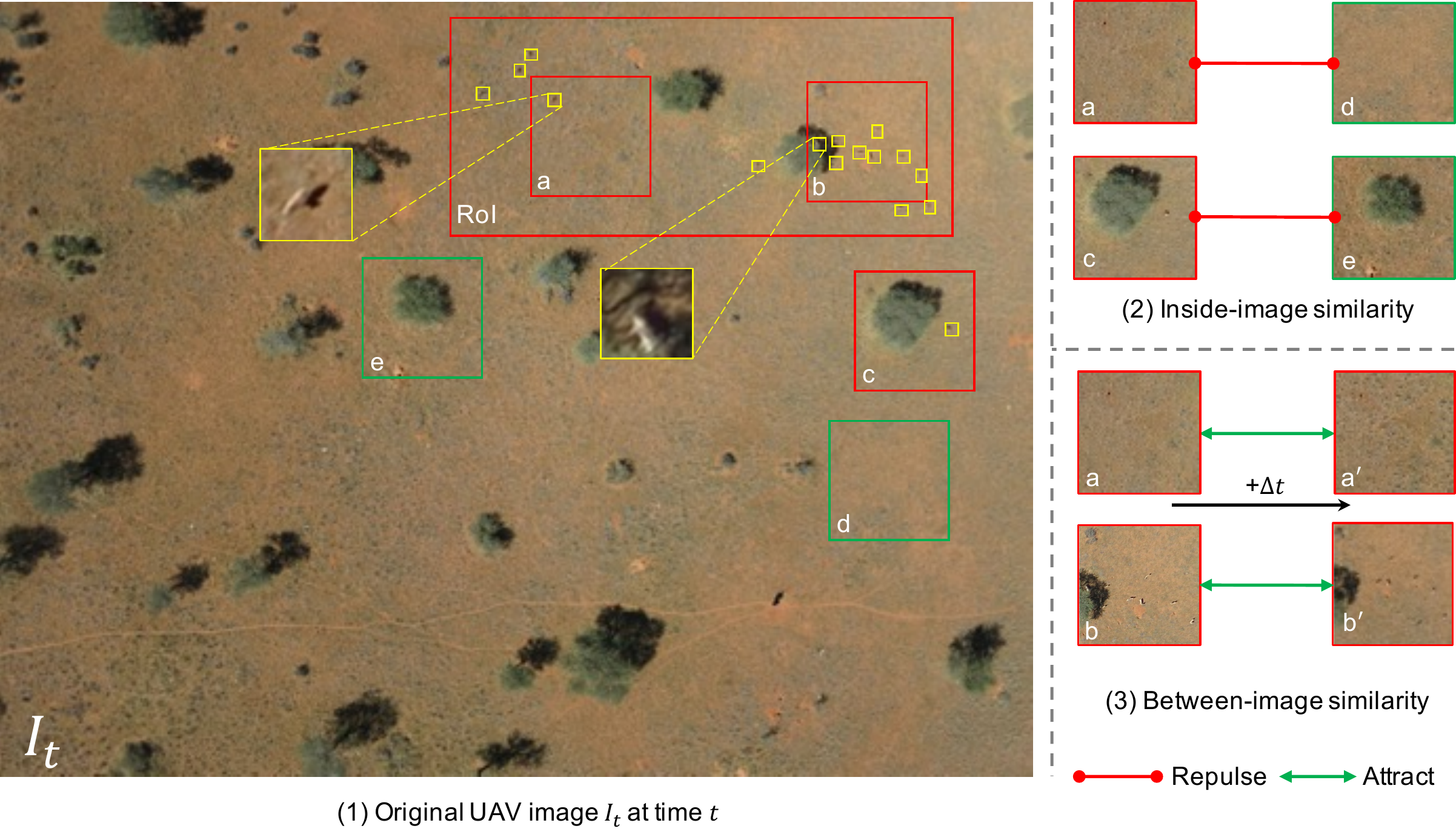}
\end{center}
\caption{Overview of Kuzikus dataset. In (1), the region of interest (RoI) (\textcolor[RGB]{255,0,0}{red} bounding box) only occupies a small part of original UAV image. Wildlife (\textcolor{yellow}{yellow} bounding boxes) is tiny which is hard to be recognized and labeled. That makes supervised learning difficult. Foreground (wildlife)/background crops are marked with \textcolor{red}{red}/\textcolor{green}{green} bounding boxes. Patches $a, c, d, e$ are cropped from different locations of the same image. Patches $a, a^\prime, b, b^\prime$ are cropped from same locations of different images (sampling time interval is $\Delta t$).}
\label{fig:correlation}
\end{figure}

\noindent\textbf{Technical Challenges.} Most contrastive models are trained on curated dataset with unique characteristics, \emph{e.g.}, ImageNet~\cite{imagenet_cvpr09}. In these datasets, images contain only a single object which is located in the center of the image (\emph{object-centric}). And objects have \emph{discriminative} visual features. The datasets also have uniformly distributed classes. In contrast, domain-specific datasets (\emph{e.g.}, our KWD) contain less discriminative visual feature, making it hard to distinguish between similar objects \emph{e.g.}, trees, wildlife from the top view. As illustrated in Figure~\ref{fig:correlation} (2), similar patches containing wildlife should be distinguished (repulsed) from the ones without wildlife. Whereas in Figure~\ref{fig:correlation} (3) similar patches from the same category should be attracted. Inside-image similarity and between-image similarity are difficult for supervised learning and contrastive learning, respectively. 
Meanwhile, the tiny objects contained in the drone image increase the difficulty of recognition. Models should be sensitive to tiny objects. The patches containing the target object account for a minimal number, which also brings difficulty to model training.


\section{Materials}

\subsection{Study Area and Data.} 

In this work, we use the data from~\cite{Beni_2018_RSE}, consisting of RGB aerial images acquired with a SenseFly eBee\footnote{\url{https://www.sensefly.com}} UAV over the Kuzikus Wildlife Reserve in Namibia\footnote{\url{https://kuzikus-namibia.de}} by the SAVMAP consortium\footnote{\url{http://lasig.epfl.ch/savmap}}. The UAV's flight height varied between 120 and 160m, resulting in a resolution of 4 to 8 cm with the given camera (Canon PowerShot S110). The images were annotated with bounding boxes for animals in a crowdsourcing campaign led by MicroMappers\footnote{\url{https://micromappers.wordpress.com}}; these annotations were then refined in several iterations by the authors. This resulted in a total of 1183 animals. We derived the \textbf{K}uzikus \textbf{W}ildlife \textbf{D}ataset \textbf{Pre}-training (KWD-Pre) and \textbf{K}uzikus \textbf{W}ildlife \textbf{D}ataset \textbf{L}ong-\textbf{T}ail distributed (KWD-LT) for pre-traning and fine-tuning/downstream task.

\subsection{Datasets} 
\noindent\textbf{KWD-Pre.} We apply the same patches creating procedure as described in~\cite{Beni_2018_RSE}. We randomly crop 4 patches for every original $4000\times3000$ image. The size of each patch is $256\times256$ pixels to save memory and have a larger batch size. We perform overlapping cropping if one image contains animals. Cropping this way increases the chances of extracting patches containing animals for training, but we do not retain any labeled information nor bounding box location. As this can be seen as a form of weak supervision in the patch extraction process, we do not know whether each patch contains animal(s) or not while applying random cropping. So the prior knowledge of classes and locations is not exploited by self-supervised learning. 

\noindent\textbf{KWD-LT.} The original images are taken by UAVs on different dates and times~\cite{Beni_2018_RSE}. We first split the original data into train, test, and validation set with a ratio of 8:1:1. Then, for the background class, we apply a random cropping procedure ($512\times512$ pixels) to the original images and verify each patch to make sure it contains no animal. For the foreground (wildlife) class, we apply a random cropping procedure ($224\times224$ pixels) around the ground truth bounding boxes to make sure each patch contains whole animal(s) body. We choose three different random seed to random cropping procedures of train, test, and validation set to make sure the cropping position is different. The train set is class imbalanced and long-tail distributed with a foreground-to-background ratio of $\frac{1}{18}$. The test and validation set are class balanced. In the experiments below, we evaluate fine-tuning on KWD-LT with different percentages of annotated animals to investigate the benefit of SSL for reduced annotation efforts.

\begin{figure}[t]
\begin{center}
  \includegraphics[width=0.5\linewidth]{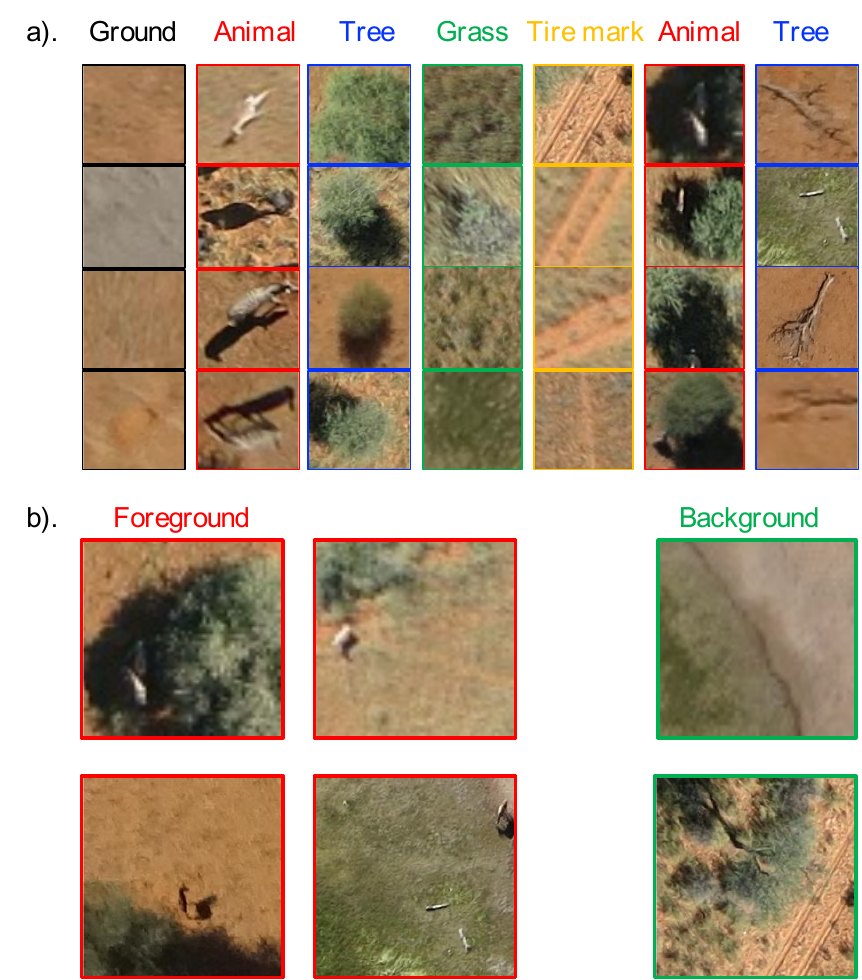}
\end{center}
  \caption{Overview of KWD-LT dataset. (a) denotes all possible elements in the dataset. Intuitively, all examples in the KWD-LT dataset consist of randomly combined elements in (a). (b) is examples of KWD-LT dataset. Instances are image-level annotated. Foreground represent images containing wildlife.}
\label{fig:dataset}
\end{figure}

\noindent\textbf{Datasets Overview.} After analyzing the content of the dataset, we can intuitively divide all elements into the following categories: ground, animal, tree, grass, and tire mark, as illustrated in Figure~\ref{fig:dataset}. Especially the animal elements in the right are animals beneath the tree and the tree elements in the right are dead tree trunks. This is extremely hard to be recognized~\cite{Beni_2018_RSE}. It is also difficult to distinguish between dead tree trunks and animal. All patches are random combination of those five elements.

\section{Methods}

\subsection{Cropping Methods}

\noindent\textbf{Random Cropping.} This method randomly selects X and Y coordinates of each patch. And based on predefined image length and width, it crops the raw images into several patches. 

\noindent\textbf{Overlapping Cropping.} Overlapping means some part of the previous patch is repeated in the current patch. Overlapping cropping is to crop raw images with stride equal to ((image length - overlapping length), (image width - overlapping width)) from left to right, top to bottom. For example, 50\% overlap means half of the previous patch is repeated in the current patch, both horizontally and vertically\footnote{For all cropping methods, we fix the random seed to improve repeatability.}.

\subsection{Contrastive Learning Framework}

\begin{figure}[t]
\begin{center}
  \includegraphics[width=0.9\linewidth]{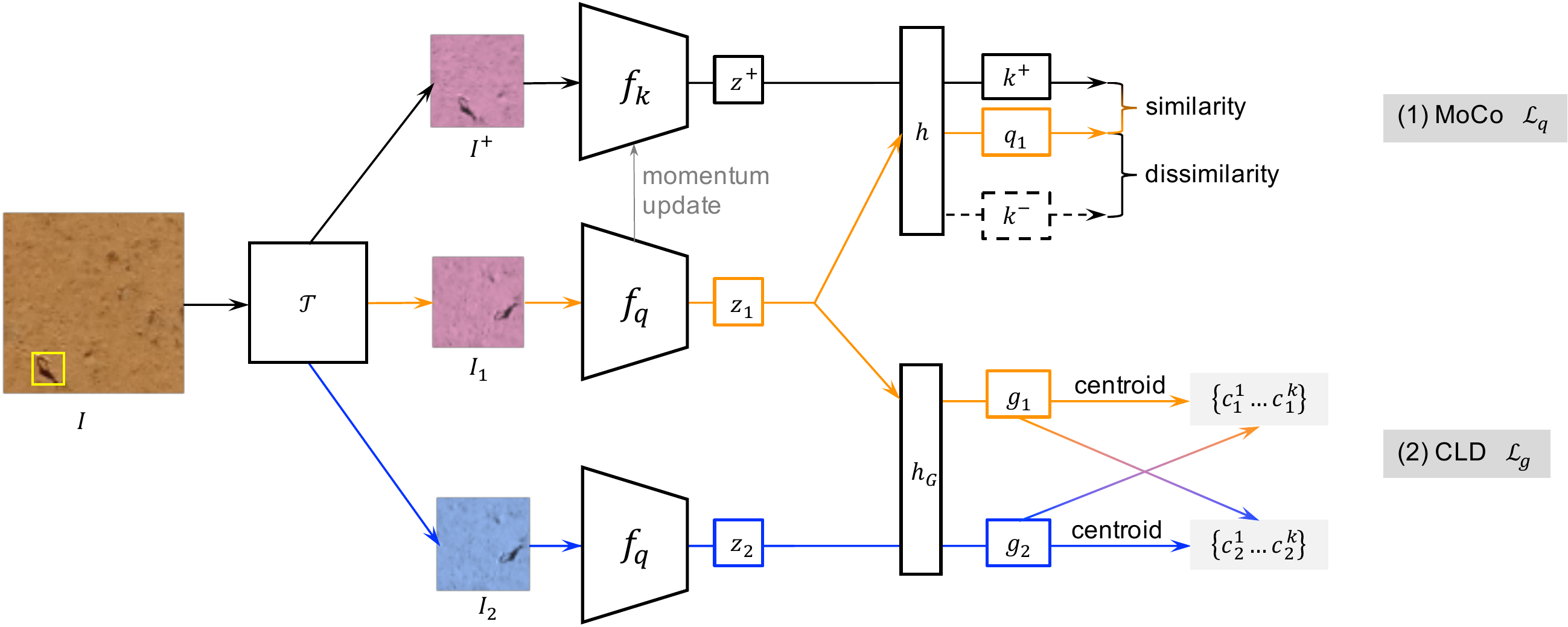}
\end{center}
  \caption{Overview of the employed SSL framework, consisting of MoCo~\cite{he2020momentum} (upper part) and CLD~\cite{Wang_2021_CVPR} (lower part). Firstly, the two different \emph{views} $I_1$ and $I^+$ of the same input $I$ are encoded by $f_q$ and $f_k$ respectively. Then, the two representations $z_1$ and $z^+$ are projected into an embedding space. $q$ and $k$ are representations of query and key in the hypersphere. Query $q_1$ and its positive key $k^+$ are from the different augmented \emph{views} of the same input and negative keys $k^-$ are encoded from different inputs (dashed bounding box). CLD first encodes two different \emph{views} $I_1, I_2$ of the same instance, then applies a different projection head and projects the representations from the same query encoders to a different embedding space from (1). Finally, a local K-Means clustering is used to find the $k$ centroids of a batch of inputs. The centroid of assigned cluster $g_1$ can be served as positive key of view $g_2$, and vice versa.}
\label{fig:model}
\end{figure}


Contrastive learning is a learning algorithm that describes similarity and dissimilarity for deep learning models. Contrastive methods train deep models to distinguish between similar and dissimilar input images in computer vision. Namely, contrastive learning uses the contrastive loss of the latent space to maximize the consistency between different augmented views of the same data point to learn representation. It constructs representations by encoding similar or different inputs.

More formally, for any data point $x$, contrastive methods aim to learn an encoder $f$ such that:
\begin{equation}
    \textbf{score} \left(f(x), f(x^+)\right) \gg \textbf{score} \left(f(x), f(x^-)\right) 
\end{equation}
where $x$ is commonly referred to as an “anchor” data point, $x^+$ is data point similar to $x$, referred to as a positive sample, $x^-$ is a data point dissimilar to $x$, referred to as a negative sample, the \textbf{score} function is a metric that measures the similarity between two features.

Contrastive methods use data augmentation to generate different \emph{views}. Positive samples are different views of the same image, while negative samples are views of different images. We choose cosine similarity as the \textbf{score} function. Given that “query”, “positive key”, and “negative keys”, which are representations of anchor, positive sample, and negative samples encoded by encoder networks, denoted as $\boldsymbol{q}$, $\boldsymbol{k}^+$, and $\boldsymbol{k}^-$, a popular choice of contrastive loss for positive pairs $(\boldsymbol{q}, \boldsymbol{k}^+)$ and negative pairs $(\boldsymbol{q}, \boldsymbol{k}^-)$ is InfoNCE~\cite{cpc}, denoted as $\mathcal{L}_q(\boldsymbol{q}, \boldsymbol{k}^+)$:
\begin{equation}
    \mathcal{L}_q(\boldsymbol{q}, \boldsymbol{k}^+)=-\log\frac{\exp\left(\boldsymbol{q}\!\cdot\!\boldsymbol{k}^+\!/\tau\right)}{\exp\left(\boldsymbol{q}\!\cdot\!\boldsymbol{k}^+\!/\tau\right)+\sum_{K} \exp\left(\boldsymbol{q}\!\cdot\! \boldsymbol{k}^-\!/\tau\right)}
\label{eq:moco}
\end{equation}
where the dictionary contains $K$ negative samples and $\tau$ denotes the temperature parameter, which is the hyper-parameter scaling the distribution of distances~\cite{cpc,he2020momentum}.


There are two methods for minimizing the empirical loss in Equation~\ref{eq:moco}. The first method is to increase the numerator $(\boldsymbol{q}\!\cdot\!\boldsymbol{k}^+)$, which means making the angle between the feature vectors of positive pairs smaller. The second method is to reduce the denominator $\sum_{K} \exp\left(\boldsymbol{q}\!\cdot\! \boldsymbol{k}^-\!/\tau\right)$, which means increasing the dissimilarity of negative pairs. Thus, the number of samples has a great influence on the quality of the learned representations~\cite{invariant}. The purpose of minimizing Equation~\ref{eq:moco} is achieved by updating the parameters of query encoder and key encoder through gradient descent.

\subsection{Momentum Contrast (MoCo)}
We apply MoCo~\cite{he2020momentum} as our unsupervised learning method. MoCo is developed based on two ideas:

\begin{itemize}
    \item \textbf{Dynamic Dictionaries.} Words in natural language processing (NLP) tasks have discrete signal spaces. Thus tokenized dictionaries, a data structure that contains all essential information units, can be constructed, which contain representations of all the words. The data in the computer vision task are distributed in high-dimensional continuous spaces, which makes it challenging to build an image dictionary. In computer vision, the dictionary should contain representations of all data points, which is difficult to implement. Thus, MoCo propose a dynamic dictionary mechanism. The dynamic dictionary is to randomly sample several images from all images during the training process and to use it as the dictionary for the current training. Each batch eliminates the earliest data encoded in the dynamic dictionary and adds the latest encoded data. This aims to keep the dictionary as consistent as possible.
    \item \textbf{Momentum Updating.} With a large dictionary, it is hard to calculate the gradient through all samples in the dictionary and to update the key encoder by back-propagation. Thus a momentum updating mechanism is proposed:
    \begin{equation}
        \theta_k \leftarrow m\theta_k + (1 - m)\theta_q
    \end{equation} where $m \in [0, 1)$ and $\theta_k, \theta_q$ are parameters of key and query encoders, respectively. Only query encoder is updated by back-propagation.
\end{itemize}

Given a batch of inputs, query and positive key are from two different augmentation \emph{views} of the same input whereas query and negative keys are from \emph{views} of different inputs. The dynamic dictionary stores both positive and negative keys. With MoCo, self-supervised learning can be regarded as a training process to perform dictionary look-up. MoCo learns image representations by matching an encoded query $q$ to a dictionary of encoded keys $k$ using a contrastive loss. Query $q$ should be similar to its matching key, positive key $k^+$, and dissimilar to negative keys $k^-$. As illustrated in Figure~\ref{fig:model}, the MoCo model consists of five parts:

(1) A \emph{stochastic data augmentation} module $\mathcal{T}$~\cite{SimCLR} transforms one given input image $I$ into different augmented \emph{views} with randomly applied augmentations, denoted $I_i$ and $I^+$ for query $q_i$ and positive key $k^+$~\cite{he2020momentum}. We sequentially apply five augmentations (Base and Color Aug. in Table~\ref{tab:aug}) similar as MoCo v2~\cite{mocov2}.
    
(2) A \emph{base encoder} $f_q$ for query $q$ maps the augmented \emph{views} into feature space: $f_q: I_i, \rightarrow \boldsymbol{z}_i,\ \text{where}\ I_i\in\mathbb{R}^{C\times W\times H}, \boldsymbol{z}_i\in\mathbb{R}^{s}$, where $z$ denotes the $s$-D encoded \emph{representation}. The parameters are updated by back-propagation. This takes the form of a CNN in our case.
    
(3) A \emph{momentum-updated encoder} $f_k$ for keys $k$ shares the same structure with \emph{base encoder} $f_q$ and is initialized with the same parameters. However, they are not learned through backpropagation during training; instead, the parameters of $f_k(\cdot)$ are updated with a momentum mechanism~\cite{he2020momentum}.
    
(4) A \emph{projection head} projects the \emph{representations} $\boldsymbol{z}_i$ into a unit hypersphere: $h: \boldsymbol{z}_i\rightarrow \boldsymbol{q}_i,\ \text{where}\ \boldsymbol{q}_i\in\mathbb{R}^{m}\ \text{and}\ \left|\left|\boldsymbol{q}_i\right|\right|=1$, the same for $\boldsymbol{z}^+$. The similarity is measured by a dot product.
    
(5) A \emph{dynamic dictionary} holds the prototypical feature for all instances~\cite{Dahua_2018_CVPR,SimCLR,Wang_2021_CVPR}.  It is implemented as a queue of fixed size, fed with the stream of mini-batches that are used for training: in current mini-batch, the encoded representations are enqueued, and the oldest are dequeued~\cite{he2020momentum}.

\begin{algorithm}[hbt!]
    \PyComment{f\_q, f\_k: encoder networks for query and key} \\
    \PyComment{queue: dictionary as a queue of K keys (CxK)} \\
    \PyComment{m: momentum} \\
    \PyComment{t: temperature} \\
    \\
    \PyCode{f\_k.params = f\_q.params} \PyComment{initialize} \\
    \\
    \PyCode{\Py{for} x \Py{in} loader:} \PyComment{load a minibatch x with N samples}\\
        \PyCode{\ \ \ \ x\_q = aug(x)} \PyComment{a randomly augmented version} \\
        \PyCode{\ \ \ \ x\_k = aug(x)} \PyComment{another randomly augmented version} \\
        \\
        \PyCode{\ \ \ \ q = f\_q.forward(x\_q)} \PyComment{queries: NxC} \\
        \PyCode{\ \ \ \ k = f\_k.forward(x\_k)} \PyComment{keys: NxC} \\
        \PyCode{\ \ \ \ k = k.detach()} \PyComment{no gradient to keys} \\
        \\
        \PyCode{\ \ \ \ \textcolor{commentcolor}{\# positive logits: Nx1}} \\
        \PyCode{\ \ \ \ l\_pos = bmm(q.view(N,1,C), k.view(N,C,1))} \\
        \\
        \PyCode{\ \ \ \ \textcolor{commentcolor}{\# negative logits: NxK}} \\
        \PyCode{\ \ \ \ l\_neg = mm(q.view(N,C), queue.view(C,K))} \\
        \\
        \PyCode{\ \ \ \ \textcolor{commentcolor}{\# logits: Nx(1+K)}} \\
        \PyCode{\ \ \ \ logits = cat([l\_pos, l\_neg], dim=1)} \\
        \\
        \PyCode{\ \ \ \ \textcolor{commentcolor}{\# contrastive loss}} \\
        \PyCode{\ \ \ \ labels = np.\Py{zeros}(N)} \PyComment{positives are the 0-th} \\
        \PyCode{\ \ \ \ loss = \Py{CrossEntropyLoss}(logits/t, labels)}\\
        \\
        \PyCode{\ \ \ \ \textcolor{commentcolor}{\# update gradients}} \\
        \PyCode{\ \ \ \ optimizer.zero\_grad()} \\
        \PyCode{\ \ \ \ loss.backward()} \\
        \PyCode{\ \ \ \ optimizer.step()} \\
        \\
        \PyCode{\ \ \ \ \textcolor{commentcolor}{\# momentum update: key network}} \\
        \PyCode{\ \ \ \ f\_k.params = m * f\_k.params + (1 - m) * f\_q.params} \\
        \\
        \PyCode{\ \ \ \ \textcolor{commentcolor}{\# update dictionary}} \\
        \PyCode{\ \ \ \ enqueue(queue, k)} \PyComment{enqueue the current minibatch} \\
        \PyCode{\ \ \ \ dequeue(queue)} \PyComment{dequeue the earliest minibatch} \\
        \\
\PyComment{bmm: batch matrix multiplication} \\
\PyComment{mm: matrix multiplication} \\
\PyComment{cat: concatenation}
    
\caption{PyTorch-like Pseudo-code of MoCo~\cite{he2020momentum}}
\label{algo:moco}
\end{algorithm}

\subsection{Clustering Based Contrastive Learning (CLD)}

\begin{figure}[t]
\begin{center}
  \includegraphics[width=0.9\linewidth]{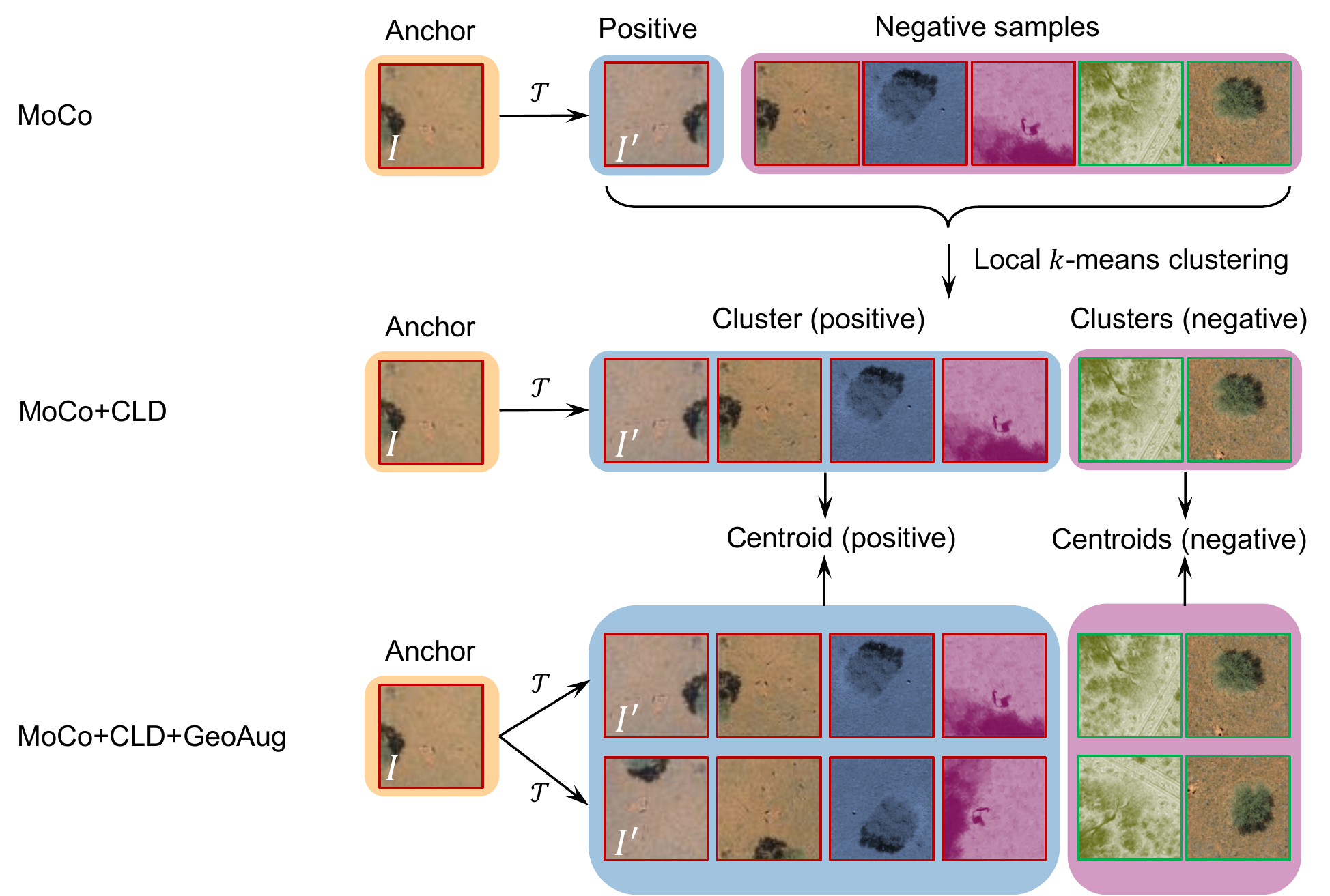}
\end{center}
  \caption{Illustration of positive and negative samples in MoCo, MoCo+CLD, and MoCo+CLD with geometric augmentation strategy scenarioes. Anchor $I$ is the original image patch on KWD dataset. $I^\prime$ is the augmented view of $I$. The \textcolor{red}{red}/\textcolor{green}{green} bounding boxes represent the foreground/background classes. In MoCo, negative samples might contain images from the same category of the anchor (positive samples). That causes the false repulsion. In MoCo+CLD, after applying local $k$-means clustering to all samples, positive and negative samples can be grouped into different categories. With CLD, there is less repulsion between samples in the same category.}
\label{fig:positive}
\end{figure}

Applying MoCo to highly correlated dataset will cause the problem of false repulsion. As shown in Figure~\ref{fig:positive}, three entries in negative samples are from the same category as the positive sample and anchor sample. Thus, those samples are repelled by MoCo. Obviously, representation learning demands encoder to generate similar features from different images in the same category. True negative samples can have a better performance than "mixing" ones in supervised learning~\cite{hard_neg_mix,hard_neg_sample}. Without access to any similarity or dissimilarity information, we aims to find a method of "sorting" positive and negative samples.

The key idea of CLD~\cite{Wang_2021_CVPR} is to cluster instances locally, and perform contrastive loss to centroids and image representaions. Therefore, similar instances are clustered into the same group and the false rejection of instances with high similarity is alleviated, as illustrated in Figure~\ref{fig:positive}.  CLD uses two different \emph{views} of the same instance as input. As such, the CLD branch shares the same query encoder $f_q$ with MoCo but uses a different projection head $h_G$, as illustrated in the lower part of Figure~\ref{fig:model}. 

To perform CLD, the unit-length features $\boldsymbol{g}_i$ of all instances in a mini-batch are first extracted from $f_q$ and $h_G$. Then, CLD implements local $k$-means clustering to $\boldsymbol{g}_i$ for a mini-batch of instances and finds $k$ local cluster centroids $\{\boldsymbol{c}_i^1,\dots,\boldsymbol{c}_i^k\}$ with $\boldsymbol{g}_i$ assigned to $\boldsymbol{C}(\boldsymbol{g}_i)$. The same operation is performed to the other branch $I_j$ from all instances in a mini-batch, denoted as $\boldsymbol{g}_j$, $\{\boldsymbol{c}_j^1,\dots,\boldsymbol{c}_j^k\}$, and $\boldsymbol{C}(\boldsymbol{g}_j)$. CLD applies the contrastive loss between $\boldsymbol{g}_i$ and clustering of the other branch $\{\boldsymbol{c}_j^1,\dots,\boldsymbol{c}_j^k\}$. Each cluster contains highly similar instances, and the assigned centroids together with representations from the other branch can be regarded as positive pairs. Namely, the centroids of the \emph{other} clusters act as negative samples~\cite{Wang_2021_CVPR}. Thus, feature vector $\boldsymbol{g}_i$ and its counterparts $\boldsymbol{g}_j$ assigned centroid $\boldsymbol{C}(\boldsymbol{g}_j)$ comprise positive pairs and all \emph{other} centroids comprise negative pairs. The local contrastive loss for CLD is

\begin{equation}
    \mathcal{L}_g(\boldsymbol{g}_i, \boldsymbol{C}(\boldsymbol{g}_j))=-\log\frac{\exp\left(\boldsymbol{g}_i\!\cdot\!\boldsymbol{C}(\boldsymbol{g}_j)\!/\tau\right)}{\sum_{\left\{\boldsymbol{c}_j^k\right\}} \exp\left(\boldsymbol{g}_i\!\cdot\! \boldsymbol{c}_j^k\!/\tau\right)}
\label{eq:cld}
\end{equation} where $\left\{\boldsymbol{c}_j^k\right\}$ denotes the set of $k$ centroids from the other branch. Thus, the loss of a dual-branch CLD in Figure~\ref{fig:model} is:

\begin{equation}
    \mathcal{L}_g(\boldsymbol{g}_1, \boldsymbol{C}(\boldsymbol{g}_2)) +  \mathcal{L}_g(\boldsymbol{g}_2, \boldsymbol{C}(\boldsymbol{g}_1))
\end{equation}
Algorithm~\ref{algo:cld} provides the pseudo-code of CLD which can be combined with classic MoCo model, described in Agorithm~\ref{algo:moco}.

We combine CLD~\cite{Wang_2021_CVPR} with MoCo v2~\cite{mocov2} and construct a total CLD contrastive loss over \emph{views} $I_1$, $I_2$, and $I^+$ with CLD weight $\lambda$ in a mini-batch. We apply different temperatures $\tau_q$ and $\tau_g$ for instance and group branches, respectively. The total CLD contrastive loss is~\cite{Wang_2021_CVPR}:
\begin{equation}
\begin{aligned}
    \mathcal{L}_{CLD} = &\frac{1}{2} \left[\mathcal{L}_q(\boldsymbol{q}_1, \boldsymbol{k}^+) + \mathcal{L}_q(\boldsymbol{q}_2, \boldsymbol{k}^+)\right]\\
    &+ \lambda\times\frac{1}{2}\left[\mathcal{L}_g(\boldsymbol{g}_1, \boldsymbol{C}(\boldsymbol{g}_2)) +  \mathcal{L}_g(\boldsymbol{g}_2, \boldsymbol{C}(\boldsymbol{g}_1)\right]
\end{aligned}
\end{equation} where $\boldsymbol{q}_{\{1,2\}}$ and $\boldsymbol{g}_{\{1,2\}}$ are feature representations issued from augmented versions of the original samples, generated following the procedure described in Section~\ref{sec:aug}.

\begin{algorithm}[hbt!]
    \PyComment{f: backbone encoder + projection mlp} \\
    \PyComment{k: k means algorithm function, outputs are the label of assigned cluster and centroids of G clusters} \\
    \\
    \PyCode{\Py{for} x \Py{in} loader:} \PyComment{load a minibatch x with N samples}\\
        \PyCode{\ \ \ \ x\_1, x\_2 = aug(x), aug(x)} \PyComment{two randomly augmented versions} \\
        \\
        \PyCode{\ \ \ \ g\_1, g\_2 = f(x\_1), f(x\_2)} \PyComment{query-like features: NxC} \\
        \\
        \PyCode{\ \ \ \ \textcolor{commentcolor}{\# label, cent: N, GxC}} \\
        \PyCode{\ \ \ \ label1, cent1 = k(g\_1)} \\
        \PyCode{\ \ \ \ label2, cent2 = k(g\_2)} \\
        \\
        \PyCode{\ \ \ \ \textcolor{commentcolor}{\# logits: NxG}} \\
        \PyCode{\ \ \ \ logits1 = mm(g\_1, cent2.T).div(T)} \\
        \PyCode{\ \ \ \ logits2 = mm(g\_2, cent1.T).div(T)} \\
        \\
        \PyCode{\ \ \ \ \textcolor{commentcolor}{\# contrastive loss}} \\
        \PyCode{\ \ \ \ loss1 = \Py{CrossEntropyLoss}(logits1/t, label2)}\\
        \PyCode{\ \ \ \ loss2 = \Py{CrossEntropyLoss}(logits2/t, label1)}\\
        \PyCode{\ \ \ \ loss = 0.5 * loss1 + 0.5 * loss2} \\
        \\
        \PyCode{\ \ \ \ \textcolor{commentcolor}{\# update gradients}} \\
        \PyCode{\ \ \ \ optimizer.zero\_grad()} \\
        \PyCode{\ \ \ \ loss.backward()} \\
        \PyCode{\ \ \ \ optimizer.step()} \\
        \\
\PyComment{mm: matrix multiplication} \\
\PyComment{div(T): tensor division with rounding the results towards zero}
\caption{PyTorch-like Pseudo-code of CLD~\cite{Wang_2021_CVPR}}
\label{algo:cld}
\end{algorithm}

\subsection{Geometric Augmentation Strategies}
\label{sec:aug}

\begin{table}
\begin{center}
\begin{tabular}{ll}
\specialrule{.1em}{.05em}{.05em} 
Module & PyTorch-like Augmentation \\
\hline
Base Aug. & RandomCrop(224)$^*$\\
& RandomHorizontalFlip(p=0.5)  \\
& GaussianBlur([0.1, 2.0])\\
\hline
Color Aug. & ColorJitter(0.4, 0.4, 0.4, 0.1)\\
&RandomGrayscale(p=0.2)\\
\hline
Rot. Aug. & RandomRotation()$^{**}$\\
\specialrule{.1em}{.05em}{.05em}
\end{tabular}
\end{center}
\caption{Overview of the employed random augmentation strategies. $^*$ To avoid information loss on tiny animals, we apply random crops without resizing. $^{**}$ We randomly rotate the images by $\{90^\circ$, $180^\circ$, $270^\circ\}$.}
\label{tab:aug}
\end{table}

State-of-the-art contrastive learning~\cite{SimCLR,simsiam,mocov2} between multiple \emph{views} of the data employs stronger augmentation strategies to improve performance. The choices of different \emph{views} have a marked impact on the performance of self-supervised pretraining~\cite{mocov2,what,LooC,tian2020view}. For different branches in CLD, \emph{e.g.}, $I_1$ and $I_2$ in Figure~\ref{fig:model}, we apply multiple augmentations to the same input image~\cite{LooC}. We keep each branch invariant to one specific augmentation transformation. For example, $I_1$ and $I^+$ are always augmented by the same color but different rotation augmentation while $I_2$ and $I^+$ are always augmented by same rotation but different color augmentation. Augmentation parameters are sampled randomly and independently from the stochastic augmentation module $\mathcal{T}$ as outlined in Table~\ref{tab:aug}. We project the queries and key into one embedding space, keeping the embedding space invariant to all augmentations. We aim to add extra geometric transformations on top of the CLD framework. Given the queries $q_1$, $q_2$ of $I_1$, $I_2$, and positive key $k^+$ of $I^+$, the loss of MoCo with our proposed geometric augmentation strategies is:
\begin{equation}
    \mathcal{L}_{Geo} = \gamma \mathcal{L}_q(q_1,k^+) + (1-\gamma)\mathcal{L}_q(q_2,k^+)
\label{geoloss}
\end{equation} where $I_1$ and $I_2$ are two different views and $\gamma$ is the weight for the combination. $\gamma$ is used to balance the color augmentation branch and geometric augmentation branch.

\subsection{Image Mixture Strategies}

\begin{figure}
     \centering
     \begin{subfigure}[b]{0.46\textwidth}
         \centering
         \includegraphics[width=\textwidth]{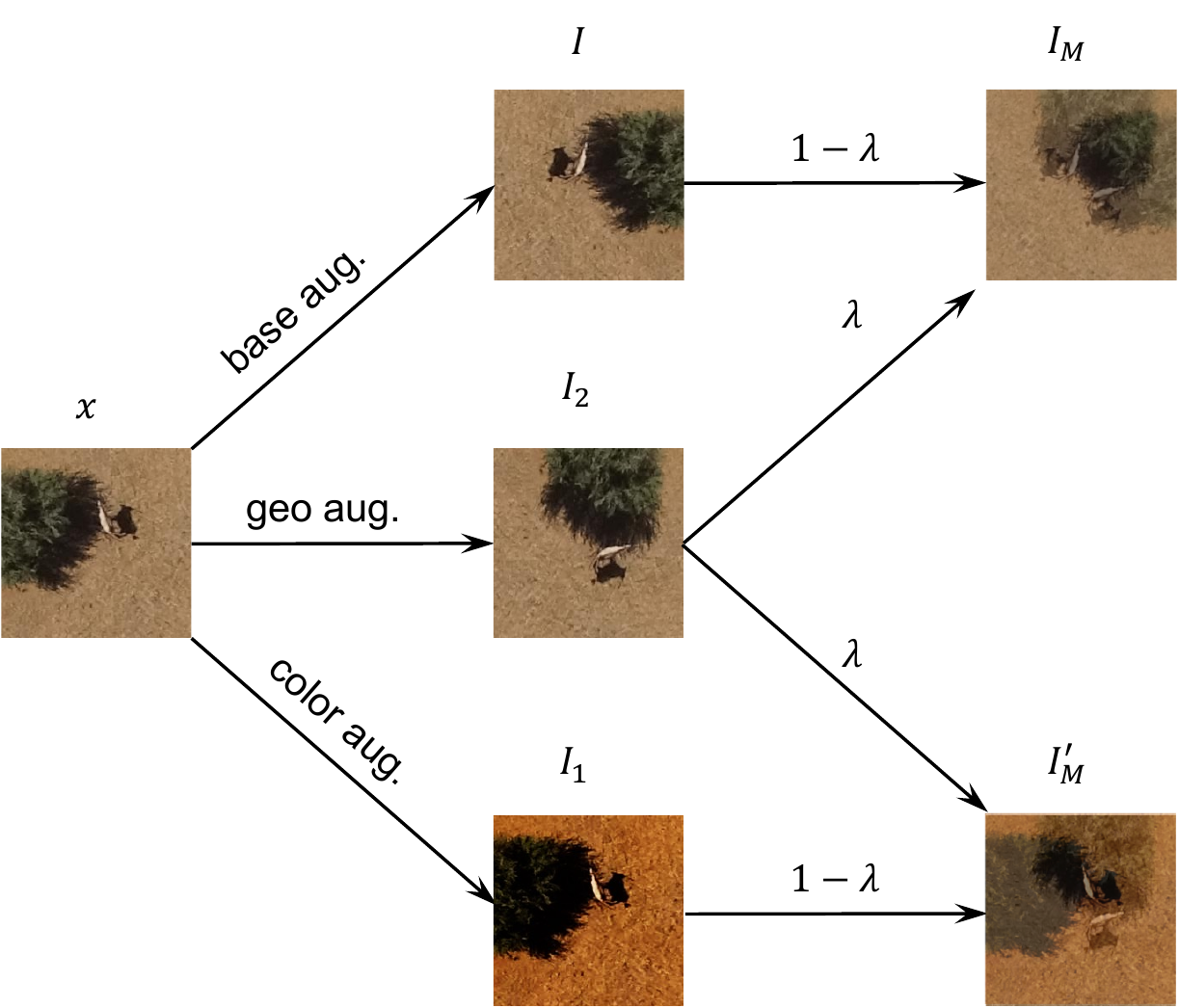}
         \caption{Views generation and image combination.}
         \label{fig:mixup1}
     \end{subfigure}
     \hfill
      \begin{subfigure}[b]{0.46\textwidth}
         \centering
         \includegraphics[width=\textwidth]{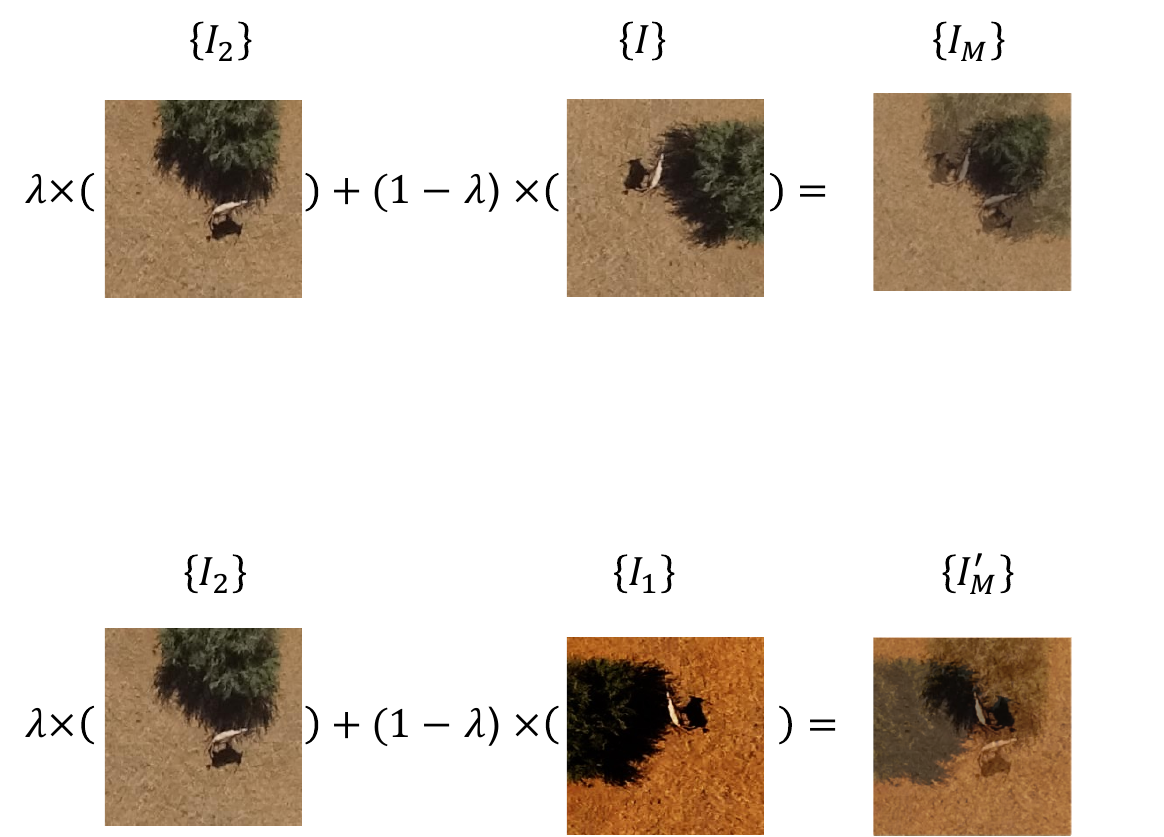}
         \caption{Illustration of mixup strategies.}
         \label{fig:mixup2}
     \end{subfigure}
\caption{Batch-wise mixing up. $I_1, I_2$ are from different augmented views of the same input.}
\label{fig:mixup}
\end{figure}

Mixup~\cite{zhang2018mixup} is a pixel-wise image-level mixture method. It combines two global images with certain weight (or mixing ratio) $\lambda$. The mixed image stores all information from two original images:
\begin{equation}
    I_M \leftarrow \lambda I_1 + (1 - \lambda) I_2
\end{equation}
where${I_M, I_1, I_2}$ denote the output mixed image and two original images, respectively.

In our method, instead of mixing two different images, we aim to combine two different views of the same input on image level. Intuitively, mixing two versions with different geometric transformation from the same foreground sample will increase the amount of target objects, as illustrated in Figure~\ref{fig:mixup}. The mixing ratio of the two augmented images and the probability of applying mixture strategy are the core information in our self-supervised model. Here we introduce two simple strategies to retain this information for loss calculation.

(1) Given $I$, $I_1$, and $I_2$ which are three different views of the same input augmented by base, color, and rotation transformations in Table~\ref{tab:aug}, we mix two images with a weight $\lambda$, as illustrated in Figure~\ref{fig:mixup}. The weight $\lambda$ is sampled from a Beta distribution $\text{Beta}(\alpha, \beta)$~\cite{shen2020mix}:
\begin{equation}
\begin{aligned}
    I_M  = \lambda I_2 + (1-\lambda) I \\
    I_M^\prime = \lambda I_2 + (1-\lambda) I_1
\end{aligned}
\end{equation}
Given that the feature vectors of $I_M$ and $I_M^\prime$ are $q_M$ and  $q_M^\prime$, we defined a \emph{mixture} loss over mixtures $I_M$ and $I_M^\prime$ as:
\begin{equation}
    \mathcal{L}_M = \lambda \mathcal{L}_q(q_M,k^+) + (1-\lambda)\mathcal{L}_q(q_M^\prime,k^+)
\label{mixtureloss}
\end{equation}
(2) We replace the MoCo loss in Equation~\ref{geoloss} by \emph{mixture} loss in Equation~\ref{mixtureloss} with a probablity of $p$. $p$ is the hyperparameter.

Obviously, for $\lambda = 1$, $\mathcal{L}_M = \mathcal{L}_q(q_1, k^+)$, while for $\lambda = 0$, $\mathcal{L}_M = \mathcal{L}_q(q_2, k^+)$, which are \emph{original} loss of MoCo v2 and MoCo with geometric augmentation, respectively.

\begin{algorithm}[hbt!]

    \PyComment{p: probability of applying mixtures} \\
    \PyComment{g: ratio of MoCo v2 loss} \\
    \PyComment{beta: hyperparameter for Beta distribution} \\
    \PyComment{lam: mixture ratio in geo-geo mixture or
     geo-color mixture}\\
    \PyComment{base\_aug: basic augmentation} \\
    \PyComment{geo\_aug: geometric augmentation} \\
    \PyComment{color\_aug: color augmentation} \\
    \\
    \PyCode{args.beta = 1.0} \\
    \\
    \PyCode{\Py{for} x \Py{in} loader:} \PyComment{load a minibatch x with N samples} \\
    \PyCode{\ \ \ \ x = base\_aug(x)} \\
    \PyCode{\ \ \ \ x\_1 = color\_aug(x)} \\
    \PyCode{\ \ \ \ x\_2 = geo\_aug(x)} \\
    \\
    \PyCode{\ \ \ \ prob = np.random.\Py{rand}(1)} \\
    \PyCode{\ \ \ \ lam = np.random.\Py{beta}(args.beta, args.beta)} \\
    \\
    \PyCode{\ \ \ \ \Py{if} prob < args.p}:
        \PyComment{Probability of applying mixtures} \\
        \PyCode{\ \ \ \ \ \ \ \ ggm = lam * x\_2 + (1 - lam) * x} \PyComment{geo-geo mixtures} \\
        \PyCode{\ \ \ \ \ \ \ \ gcm = lam * x\_2 + (1 - lam) * x\_1} \PyComment{geo-color mixtures} \\
        \PyCode{\ \ \ \ \ \ \ \ \textcolor{commentcolor}{\# if choosing mixture strategy, we only compute the }} \\
        \PyCode{\ \ \ \ \ \ \ \ \ \ \textcolor{commentcolor}{'mixture' loss instead of the origin loss.}} \\
        \PyCode{\ \ \ \ \ \ \ \ loss = lam * model(ggm, x) + (1 - lam) * model(gcm, x)} \\ 
    \PyCode{\ \ \ \ \Py{else}:} \\
    \PyCode{\ \ \ \ \ \ \ \ \textcolor{commentcolor}{\# compute origin loss}} \\
        \PyCode{\ \ \ \ \ \ \ \ loss = g * model(x\_1, x) + (1 - g) * model(x\_2, x)} \\
        \\
    \PyCode{\ \ \ \ \textcolor{commentcolor}{\# update gradients}} \\ 
    \PyCode{\ \ \ \ optimizer.zero\_grad()} \\
    \PyCode{\ \ \ \ loss.backward()} \\
    \PyCode{\ \ \ \ optimizer.step()}
\caption{PyTorch-like Pseudo-code of Our Mixture Strategy.}
\label{algo:mixup}
\end{algorithm}



\subsection{Models Design and Comparison}

\begin{table}
\begin{center}
\begin{tabular}{ll|c|c|c}
\specialrule{.1em}{.05em}{.05em} 
Name & Backbone & CLD & Geometric Strategies & Mixture Strategies\\
\hline
MoCo v2~\cite{he2020momentum,mocov2} & MoCo v2 &  &  \\
MoCo+CLD~\cite{Wang_2021_CVPR} & MoCo v2 & \checkmark  & \\
MoCo Geo & MoCo v2 & & \checkmark & \\
GeoCLD & MoCo v2 & \checkmark & \checkmark & \\
MixCo & MoCo v2 &  & \checkmark & \checkmark \\
\specialrule{.1em}{.05em}{.05em}
\end{tabular}
\end{center}
\caption{MoCo based models.}
\label{table: models}
\end{table}

In this work, we apply MoCo v2~\cite{mocov2} as our contrastive baseline model. We explore the performance of MoCo v2~\cite{mocov2} and MoCo+CLD~\cite{Wang_2021_CVPR} on KWD datasets with self-supervised pretraining. \textbf{MoCo Geo} replace the augmentation method of MoCo v2 with geometric augmentation in Table~\ref{tab:aug}. Based on the methods we proposed, we construct two models: MoCo+CLD with geometric augmentation strategies (\textbf{GeoCLD}) and MoCo with mixture augmentation strategies (\textbf{MixCo}). We compare those MoCo based models with supervised models. For supervised models, we use a ResNet-50~\cite{resnet}, pretrained on ImageNet~\cite{imagenet_cvpr09}. \textbf{Sup1} freezes the output of ResNet-50 average pooling layer. \textbf{Sup2} fine-tunes the pretrained ResNet-50 on KWD-LT with full labels, as shown in Figure~\ref{table: models}.  denoted as MCC0. Instead of using a RandomResizeCrop, we apply a PyTorch RondomCrop to input images to keep more information. We set the $\lambda$, number of clusters to 0.25 and 32 respectively. For \textbf{MixCo} we set $\gamma = 0.9$ and $p = 0.3$. Detailed information of different models are outlined in Table~\ref{table: models}.

\begin{figure}
     \centering
     \begin{subfigure}[b]{0.49\textwidth}
         \centering
         \includegraphics[width=0.9\textwidth]{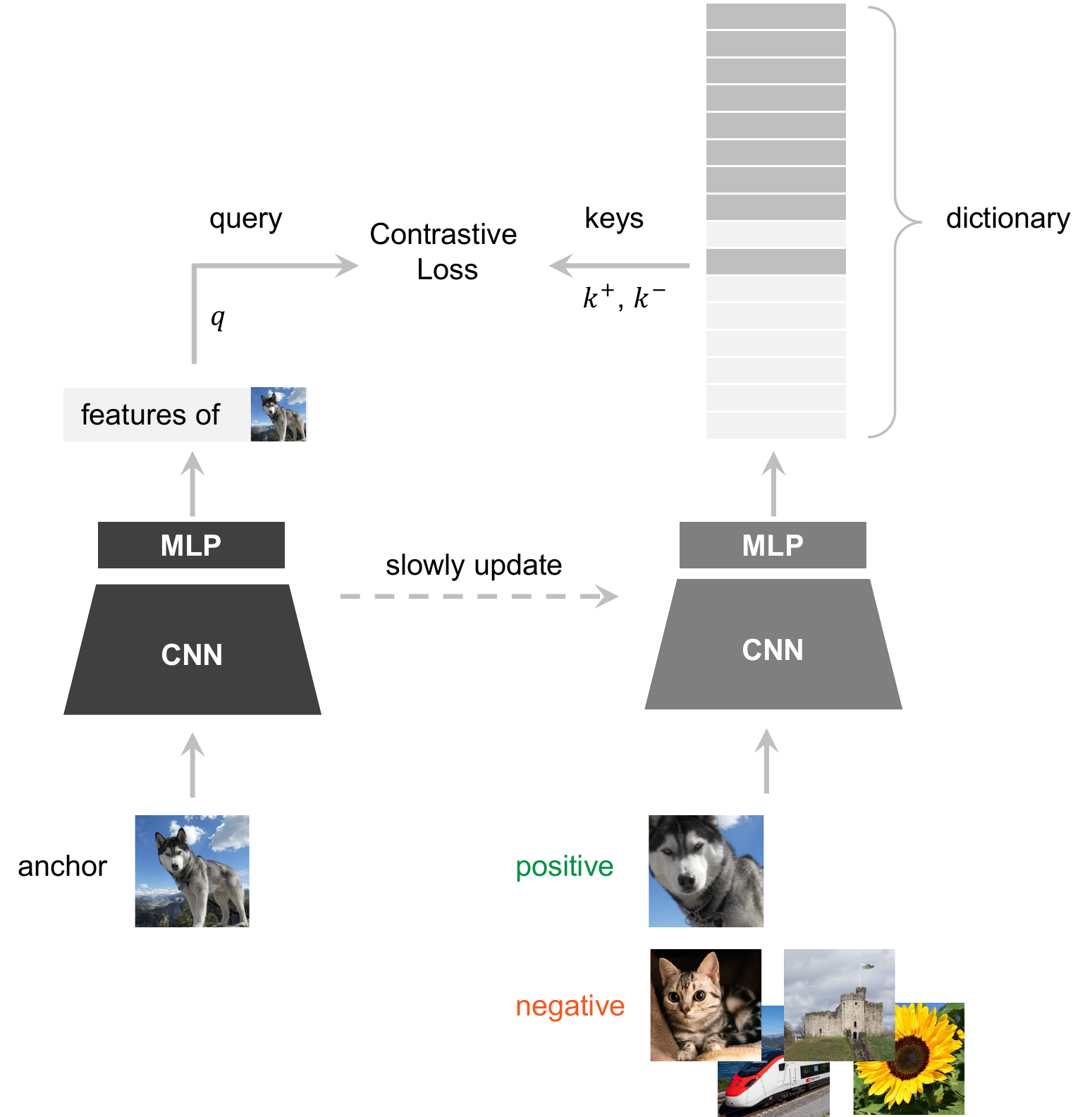}
         \caption{Original MoCo~\cite{he2020momentum}.}
         \label{fig:mocov2}
     \end{subfigure}
     \hfill
      \begin{subfigure}[b]{0.49\textwidth}
         \centering
         \includegraphics[width=0.6\textwidth]{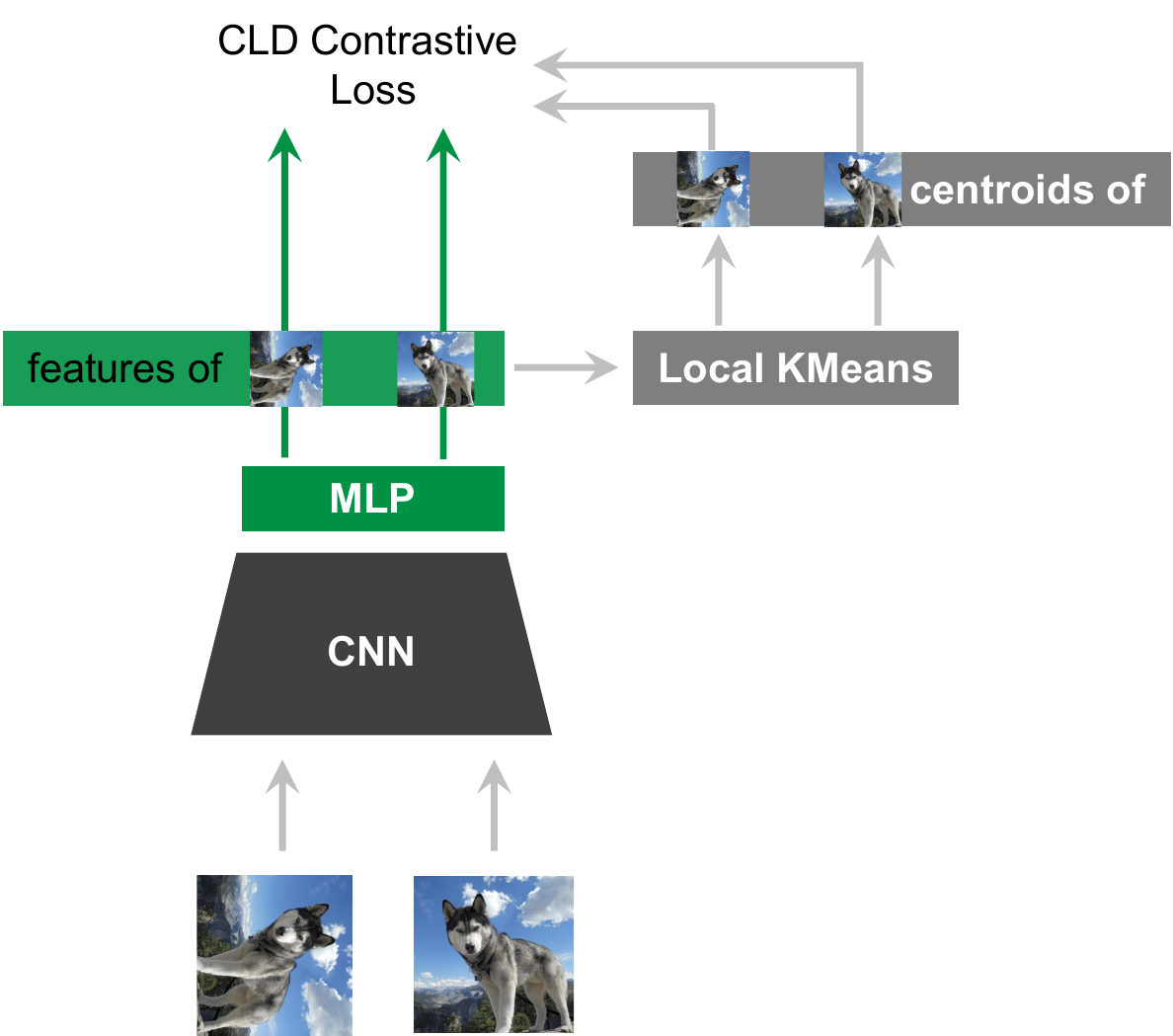}
         \caption{Original CLD~\cite{Wang_2021_CVPR}}
         \label{fig:cld}
     \end{subfigure}
     \hfill
     \begin{subfigure}[b]{0.49\textwidth}
         \centering
         \includegraphics[width=\textwidth]{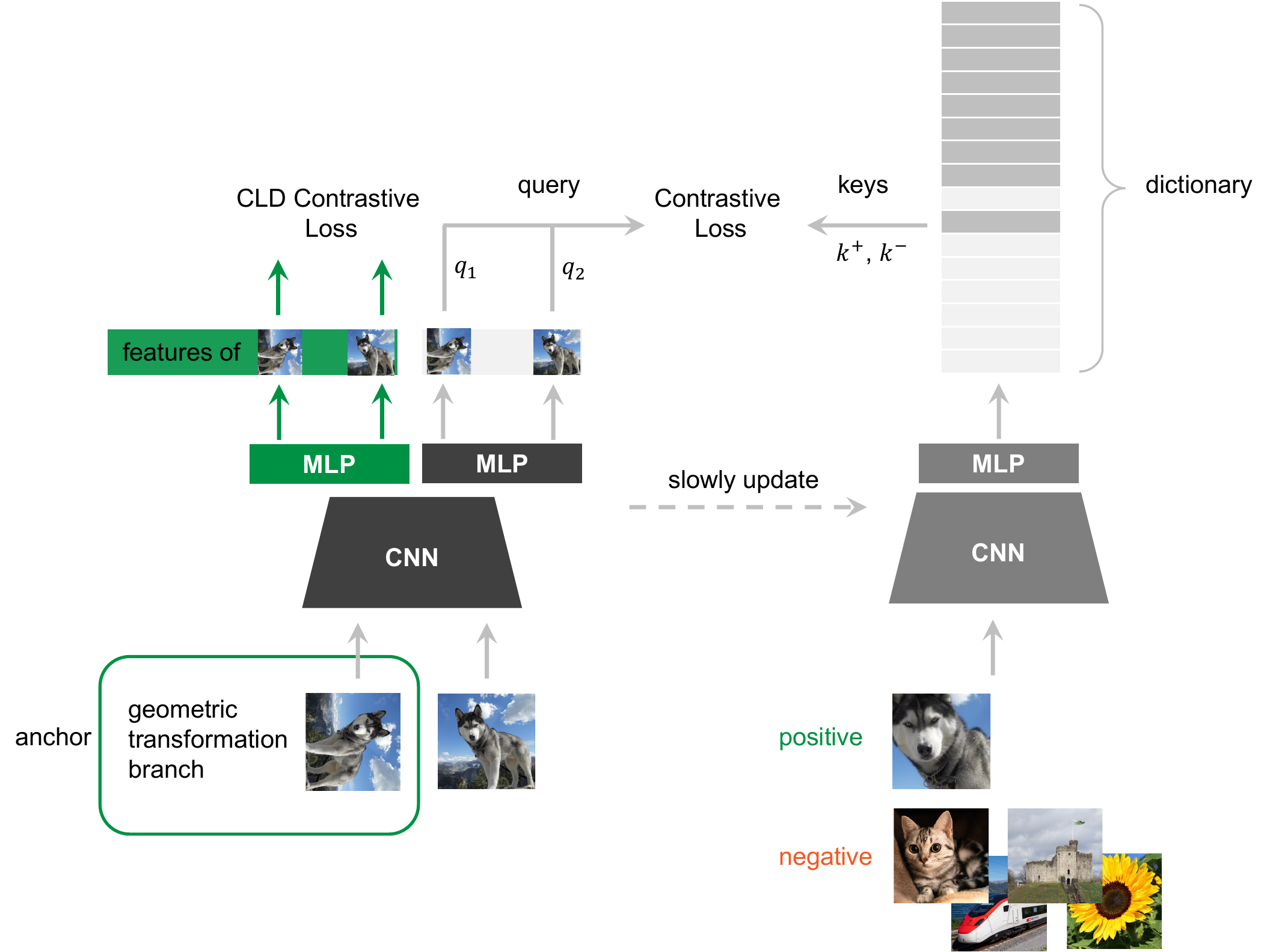}
         \caption{GeoCLD.}
         \label{fig:geocld}
     \end{subfigure}
     \hfill
     \begin{subfigure}[b]{0.49\textwidth}
         \centering
         \includegraphics[width=0.95\textwidth]{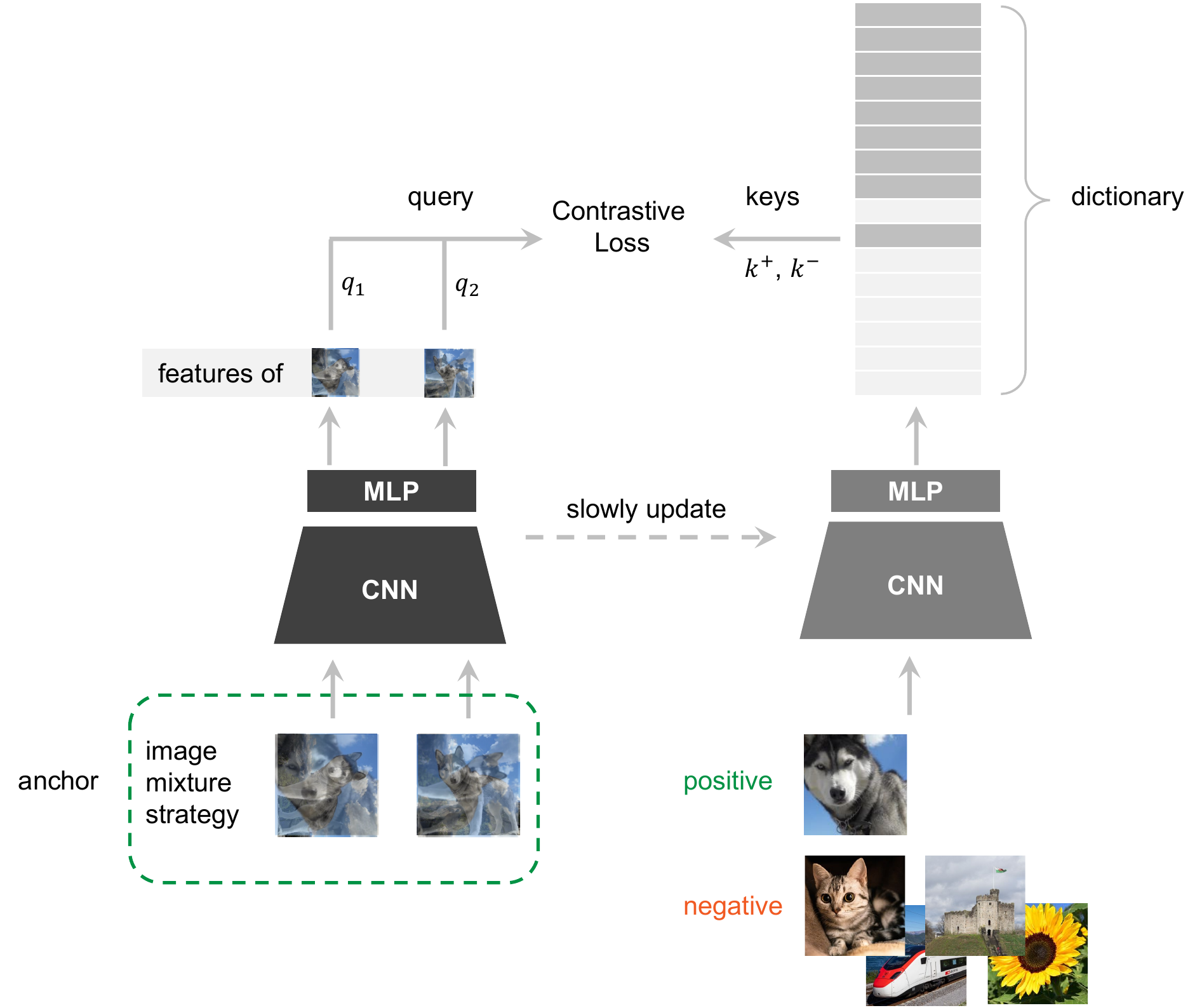}
         \caption{MixCo.}
         \label{fig:mixco}
     \end{subfigure}
\caption{Overview and comparison of MoCo, CLD, GeoCLD, and MixCo}
\label{fig:modelscompare}
\end{figure}

Here is the conceptual comparison of four model mechanisms, as illustrated in Figure~\ref{fig:modelscompare}. (a) MoCo has two branches: query branch and key branch. MoCo encodes the latest keys by a momentum-updated key encoder and maintains a dictionary of keys. (b) CLD head has two symmetric branches and two views of inputs. Two symmetric branches are used to compute centroids of a \emph{mini}-batch of representations. CLD incorporates local clustering into an instance discrimination \emph{pretext} task. The centroid of the assigned cluster serves as a positive key and the other centroids as negative keys. The CLD head is trained without a memory bank in a contrastive way like SimCLR~\cite{SimCLR}. (c) \textbf{GeoCLD} is a combination of MoCo and CLD. The MoCo and CLD share the same inputs and encoder but use a different projection head. We apply geometric augmentation to one group of inputs. (d) MixCo has the same structure as MoCo. Instead of MoCo v2 augmentation methods, we apply our mixup strategies to the inputs of MixCo and then add a symmetric branch to original MoCo model. Parameters of all models are updated by minimizing the total loss: 

\begin{equation}
\begin{aligned}
   &\textbf{GeoCLD:}\quad \mathcal{L}_{CLD} = \frac{1}{2} \left[\mathcal{L}_q(\boldsymbol{q}_1, \boldsymbol{k}^+) + \mathcal{L}_q(\boldsymbol{q}_2, \boldsymbol{k}^+)\right] + \lambda\times\frac{1}{2}\left[\mathcal{L}_g(\boldsymbol{g}_1, \boldsymbol{C}(\boldsymbol{g}_2)) +  \mathcal{L}_g(\boldsymbol{g}_2, \boldsymbol{C}(\boldsymbol{g}_1)\right] \\
   &\textbf{MixCo (with mixup):}\quad \mathcal{L}_M = \lambda \mathcal{L}_q(q_M,k^+) + (1-\lambda)\mathcal{L}_q(q_M^\prime,k^+)\\
   &\textbf{MixCo (without mixup):}\quad \mathcal{L}_{Geo} = \gamma \mathcal{L}_q(q_1,k^+) + (1-\gamma)\mathcal{L}_q(q_2,k^+)
\end{aligned}
\end{equation}

Our idea is to keep the original model framework and apply our domain-specific strategies to improve the downstream task performance.

\section{Experiments}

\noindent\textbf{Optimizer.} We use stochastic gradient descent for self-supervised pretraining and downstream task fine-tuning. For the self-supervised pretraining, we apply the same cosine decay scheduler as proposed in~\cite{Wang_2021_CVPR}. For the semi-supervised fine-tuning, the initial learning rate is 0.01 and we likewise apply a cosine decay schedule~\cite{SimCLR}. For the downstream task, we set the initial learning rate as 30 and we apply same strategy proposed in~\cite{he2020momentum}. 

\noindent\textbf{Training details.}
All models are trained on a single node with a single GPU. We pretrain our SSL model on one Nvidia GeForce RTX 3090 (with 24 GB memory) for 200 epochs with a batch size of 64. We train an ImageNet pretrained ResNet-50 end-to-end with labels on Kuzikus-Patch dataset using EPFL cluster. The linear classifier is trained on one Nvidia TITAN V (with 12 GB memory) for 100 epochs with a batch size of 256. We fine-tune the model \emph{end-to-end} on one Nvidia GeForce RTX 3090 (with 24 GB memory) for 200 epochs with a batch size of 256. One SSL pretraining costs 14-24 hours. 

\noindent\textbf{Base Encoder and Projection Head.} 
We apply a ResNet-50 without pretrained on ImageNet as our base encoder. We simply remove the last fully-connected layer and use the output of average pooling as feature vector $\boldsymbol{z}_i$. We adopt a Multi-Layer Perceptron (MLP) head following~\cite{SimCLR,mocov2}, which is a 2-layer MLP (2048-dimensional hidden layer, with ReLU). We share the hidden layer and apply a different final layer of the MLP head to MoCo branch and CLD branch. The dimension of the unit-length feature representation $q_i$ and $k^{\{+,-\}}$ is 128.

\noindent\textbf{Hyperparameter Choice.}
For fair comparison and avoiding hyper-parameters tuning redundancy, we select the CLD weight $\lambda$ and number of cluster by linear classification on frozen features with labeled data. According to our prior knowledge, the images in the KWD dataset are primarily composed of artefacts as follows: animal, tree, grass, tire mark (road), animal beneath tree, and dead tree trunk, as illustrated in Figure~\ref{fig:dataset}. Ideally, the number of all possible clusters is therefore $C_6^0 + C_6^1 + C_6^2 + C_6^3 + C_6^4 + C_6^5 + C_6^6 = 2^6 = 64$. Among all elements, animals beneath tree are hard to recognize; also dead tree trunks are easy confused with animals. Meanwhile, with a batch size of 64, it might not yield all 64 combinations, but we still use 64 clusters to give the model enough freedom. We train the MoCo+CLD model with different hyper-parameters for 200 epochs and select the best hyper-parameter combination based on accuracy on the validation set of the downstream recognition task.


\noindent\textbf{Downstream Recognition Task.} 
We verify different models by applying linear classification on encoded image representations. We follow the same common linear classification protocol as~\cite{he2020momentum}. We first perform self-supervised pretraining on KWD-Pre dataset. Then we perform two kinds of experiments: (1) \emph{frozen} features: we freeze the output features of the global average pooling layer of a ResNet and train a linear classifier (a fully-connected layer followed by softmax)~\cite{he2020momentum} in a supervised way on our KWD-LT downstream task dataset; (2) \emph{end-to-end}: we fine-tune the base encoder and linear classifier by softmax loss instead of contrastive loss. And we report the linear classification top-1 accuracy on the KWD-LT validation set, as well as recall and precision for foreground class.



\section{Results and Discussion}

\subsection{Self-Supervised Pretraining}

\begin{table}
\begin{center}
\begin{tabular}{lccc}
\specialrule{.1em}{.05em}{.05em} 
Model & Acc & Prec & Rec \\
\hline
ImageNet Pretraining (Sup1) & 86.7 & 96.5 & 76.1 \\
End-to-End Finetuning (Sup2) & 88.6 & 99.3 & 77.7 \\
\hline
MoCo v2 & 83.9 & 95.3 & 71.1 \\
MoCo+CLD& 92.8 & 98.5 & 86.8 \\
\hline
MoCo Geo \textbf{\textcolor{green!80!blue}{(ours)}}& 88.9 & 96.6 & 80.4 \\
GeoCLD \textbf{\textcolor{green!80!blue}{(ours)}} & 93.6 & 99.5 & 87.4 \\
MixCo \textbf{\textcolor{green!80!blue}{(ours)}} & 94.2 & 99.3 & 89.0 \\
\specialrule{.1em}{.05em}{.05em} 
\end{tabular}
\end{center}
\caption{\textbf{Linear classifier} top-1 accuracy (\%), foreground class precision and recall (\%) on \emph{frozen} features with 10\% labels, comparison of self-supervised learning on KWD-Pre (MoCo v2, MoCo+CLD, MoCo Geo, GeoCLD, MixCo) and supervised pretraining on ImageNet (Sup1, Sup2) which is trained with full labels.}
\label{tab:overall}
\end{table}

\begin{figure}[t]
\begin{center}
  \includegraphics[width=0.6\linewidth]{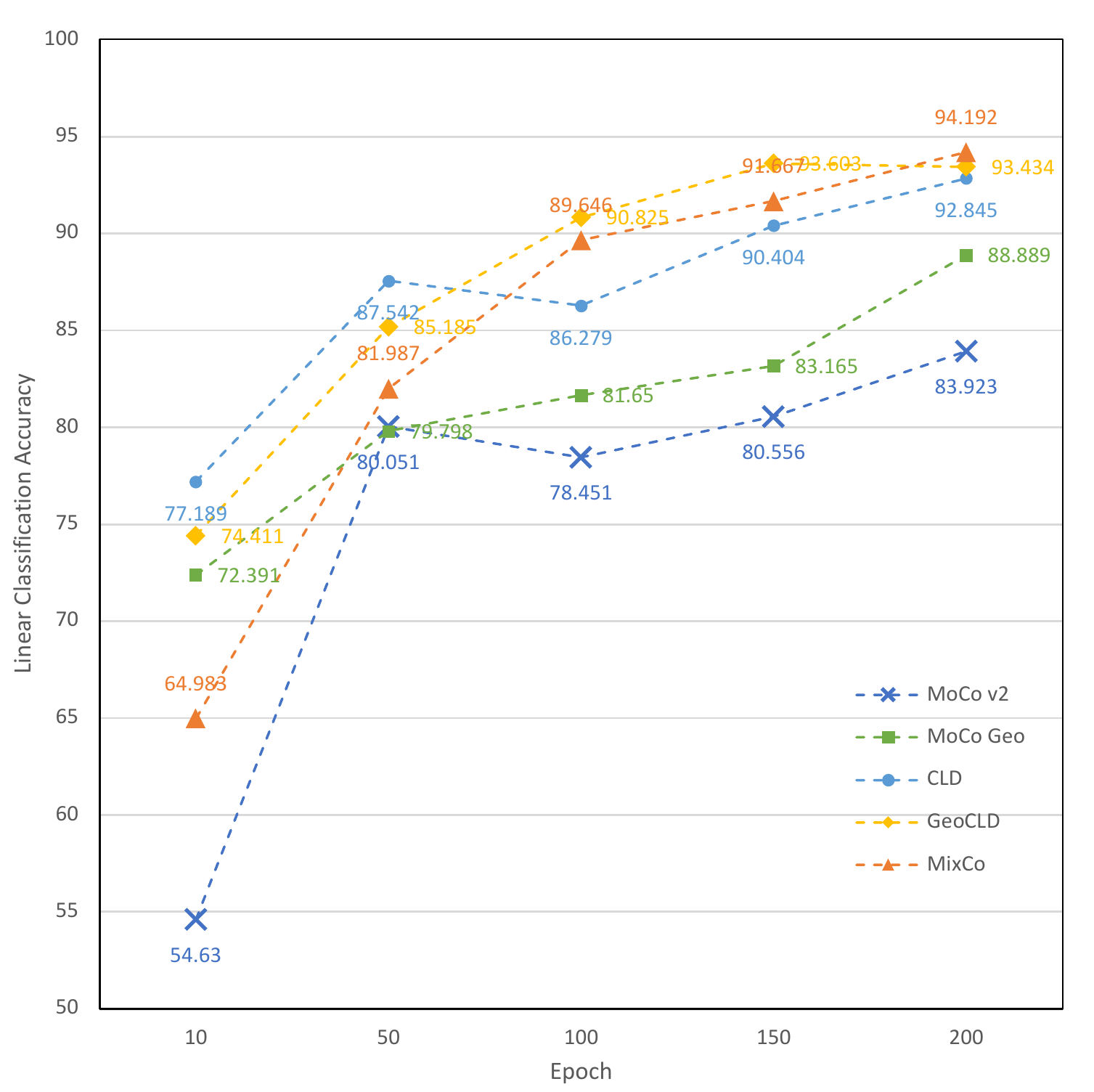}
\end{center}
  \caption{\textbf{Comparison of five contrastive loss mechanisms} under the KWD-LT linear classification protocol. We only vary the models in Table~\ref{table: models}.}
\label{fig:200epochoverall}
\end{figure}

\begin{figure}
     \centering
     \begin{subfigure}[b]{0.48\textwidth}
         \centering
         \includegraphics[width=\textwidth]{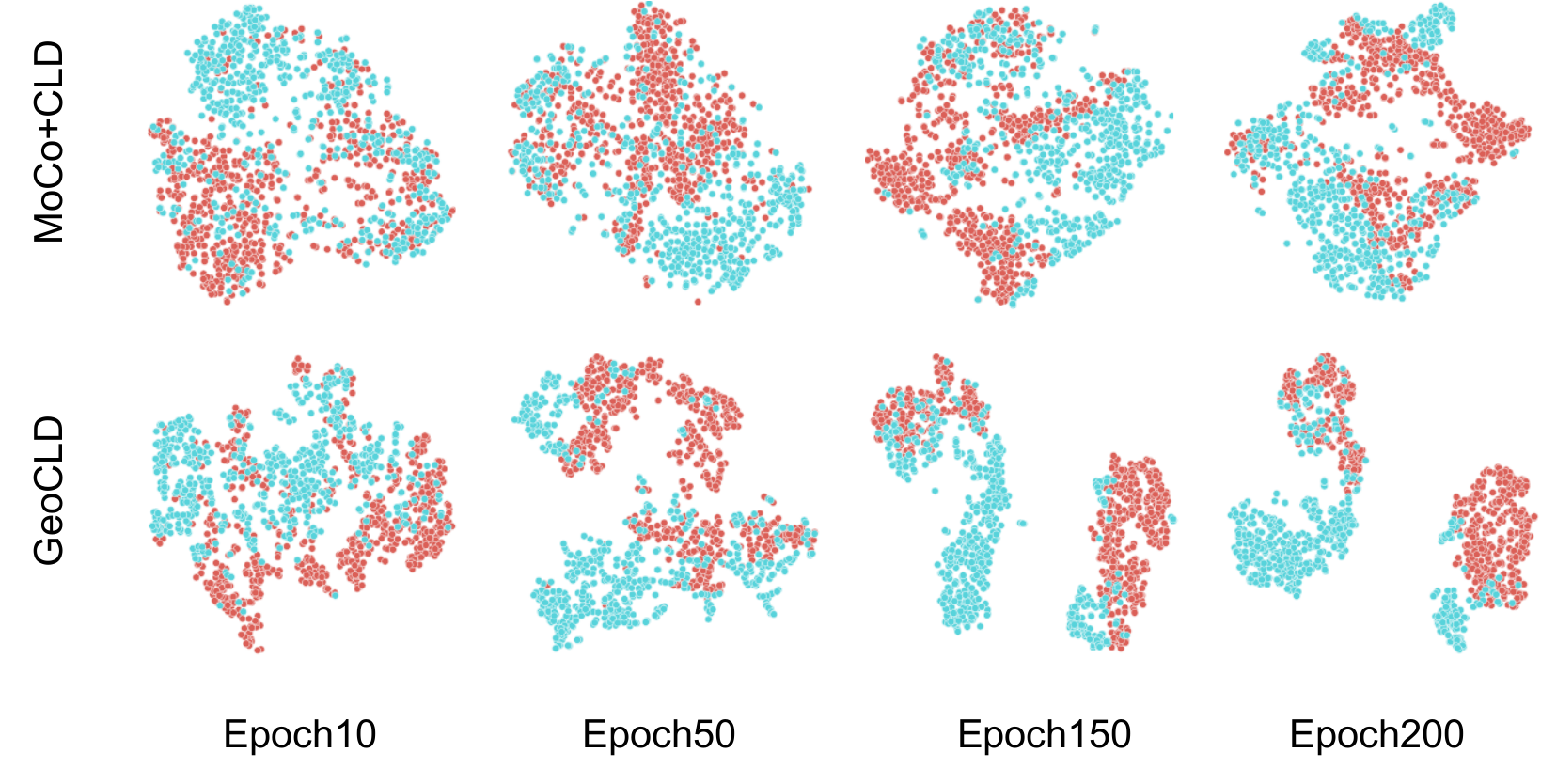}
         \caption{The loss of MoCo v2 and MixCo.}
         \label{tsne0}
     \end{subfigure}
     \hfill
     \begin{subfigure}[b]{0.48\textwidth}
         \centering
         \includegraphics[width=\textwidth]{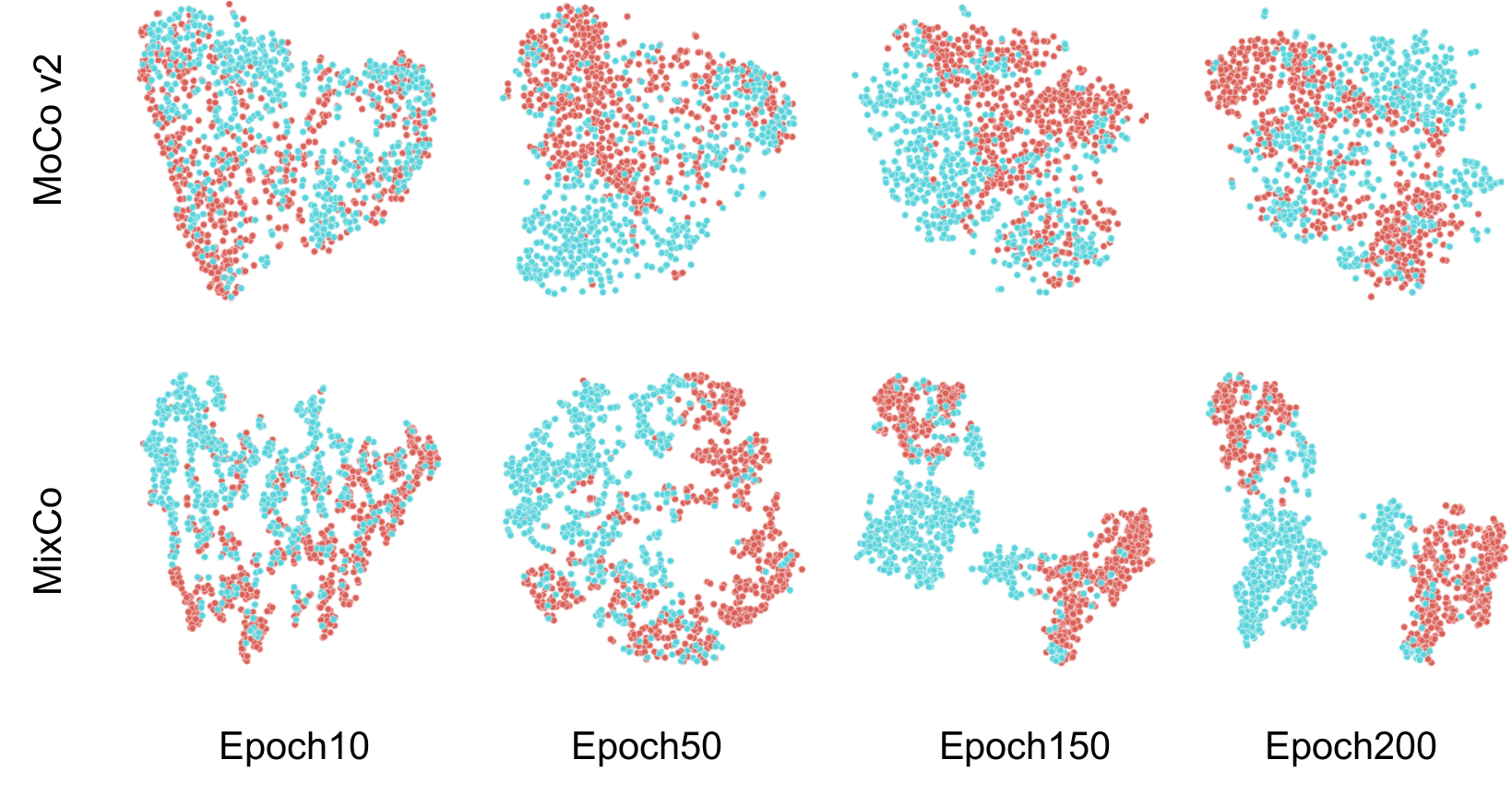}
         \caption{The KNN accuracy of MoCo v2 and MixCo.}
         \label{tsne1}
     \end{subfigure}
\caption{\textbf{t-SNE feature visualization} of MoCo v2, MoCo+CLD, GeoCLD, and MixCo on KWD-LT. All our proposed methods have earlier and better separation between foreground and background classes (indicated by color) than MoCo v2.}
\label{fig:tsne}
\end{figure}

We verify different pretraining models by linear classification accuracies. We follow the same common linear classification protocol as~\cite{he2020momentum}. We first perform self-supervised pre-training on KWD-Pre dataset. We freeze the output features of the global average pooling layer of a ResNet. Then we train a linear classifier in a supervised way on our KWD-LT dataset and report the top-1 classification accuracy on the KWD-LT validation set. For the hyper-parameters and training process of the linear classifier, we follow the same settings as \cite{he2020momentum}. 

The possible repeated and highly correlated patches slow the training process and lower the performance of SSL pretraining as it breaks the instance discrimination presumption described in Section~\ref{intro}. For a fair comparison, we evaluate the performance of self-supervised pretraining by linear classification of \emph{frozen} features with 10\% labels. The linear classifier accuracy in Table~\ref{tab:overall} shows that SSL pretraining on the target dataset is an intense competition of supervised pretraining on ImageNet. MoCo with CLD (MoCo+CLD) performs much better than supervised fine-tuning (Sup2). With controlled and designed augmentation for our scenario, GeoCLD outperforms MoCo+CLD by 0.8\%, and MixCo outperforms MoCo+CLD by 1.4\%. We can tell that SSL pretraining with controlled augmentation can improve the performance of rare wildlife recognition. Our strategies help learn geometric invariant information in the latent space as well as the less discriminative visual features in the raw images. However, in Figure~\ref{fig:200epochoverall}, the accuracy of GeoCLD in epoch 200 is slightly lower than that in epoch 150. That might imply that na\"ively adding geometric transformation might cause the problem of overfitting. The training process of MoCo fluctuates and applying geometric or image mixup strategies can stable the training and increase the accuracy steadily. Furthermore, feature visualization in Figure~\ref{fig:tsne} shows that CLD with geometric augmentation (GeoCLD) and MixCo converge faster and better towards a more distinctive feature representation than MoCo v2 and MoCo+CLD. 

\subsection{Fine-Tuning with Few Labels}

Self-supervised pretraining on our dataset can utilize annotations far more efficiently than supervised pretraining on ImageNet. As shown in Table~\ref{tab:fewshot}, fine-tuning the encoder \emph{end-to-end} will completely destroy the capacity of ImageNet pretrained model. Even though we freeze the feature representations, the model perform poorly with small fraction of labeled instances. However, fine-tuning only linear classifier on \emph{frozen} features with 1\% and 10\% annotations outperforms fine-tuning the encoder \emph{end-to-end}. When we only train the linear classifier with 1\% of labels, GeoCLD outperforms MoCo v2 by 9.4\% and MoCo+CLD by 11.5\%. For MoCo+CLD, fine-tuning encoder with 10\% annotations outperforms the recognition accuracy with full annotations. For GeoCLD, we need 20\% annotations to get a better result. Meanwhile, for MoCo+CLD, fine-tuning the encoder with 20\% of labeled instances will have the same performance with fine-tuning ImageNet pretrained model (Sup2) with full labeled instances. Whereas for GeoCLD, only 10\% labeled data is required to outperform the Sup2 supervised model. The results show that SSL model can learn more information through geometric invariant mapping and capturing geometric invariant information can benefit UAV top view imagery task. 

The results of Sup1, Sup2, MoCo v2, and MoCo+CLD in Table~\ref{tab:overall} and of these \emph{frozen} features in Table~\ref{tab:fewshot} show that increasing the fraction of labels used can improve the performance of linear classification. But it has a bottleneck performance of 88.6\%. This is different in training linear classifier on GeoCLD. Although adding extra geometric augmentation can boost performance. Feeding all annotated data can decrease the performance of \emph{frozen} features in GeoCLD, as shown in Table~\ref{tab:overall} and~\ref{tab:fewshot}. That could be caused by imbalanced classes: the model overfits to the background class and is over-confident.

\begin{table}
\begin{center}
\setlength{\tabcolsep}{4pt}
\begin{tabular}{ll|ccc|ccc|ccc}
\specialrule{.1em}{.05em}{.05em} 
 Model &Fine-tuning&\multicolumn{3}{c|}{1\% labels}& \multicolumn{3}{c|}{10\% labels}  & \multicolumn{3}{c}{20\% labels} \\
 strategies & (on KWD-LT)& Acc & Prec &Rec & Acc & Prec & Rec & Acc & Prec & Rec\\
\hline
Sup2&\emph{end-to-end}& 50.4 & 0 & 0 & 69.8 & 98.8 & 40.2 & 80.5 & 97.3 & 62.8\\
Sup1&\emph{frozen} features & 54.4 & 100.0 & 9.0 & 69.6 & 99.3 & 39.3 & 76.2 & 99.5 & 52.7\\
\hline
\multirow{2}{*}{MoCo v2}&\emph{end-to-end} & 68.0 & 98.5 & 35.9 & 76.8 & 99.7 & 53.4 & 77.9 & 99.7 & 55.4 \\
&\emph{frozen} features & 74.0 & 97.1 & 49.1 & 76.9 & 96.8 & 55.1 & 77.4 & 97.3 & 56.1\\\hline
\multirow{2}{*}{MoCo+CLD}&\emph{end-to-end} & 70.5 & 98.5 & 40.9 & 76.0 & 99.7 & 51.6 & 88.5 & 99.3 & 77.4 \\
&\emph{frozen} features & 71.9 & 94.8 & 46.0 & 82.5 & 98.2 & 66.0 & 83.8 & 98.2 & 68.8\\\hline
\multirow{2}{*}{GeoCLD}&\emph{end-to-end} & 78.9 & 98.0 & 58.7 & 90.7 & 99.8 & 81.3 &\textbf{91.9}&100&83.7\\
&\emph{frozen} features & \textbf{83.4}&97.0&68.6& \textbf{91.7}&98.8&84.5&90.1&99.1&81.4\\
\specialrule{.1em}{.05em}{.05em} 
\end{tabular}
\end{center}
\caption{\textbf{Animal recognition} accuracy when using a portion of the available labeled samples in the final classifier (\emph{frozen} features) or in the base encoder (\emph{end-to-end}) and when varying the type of pretraining, the self-supervised strategy and the type of fine tuning. Prec = Precision, Rec = Recall. $^*$ The results are not updated and are different from Table~\ref{tab:overall} because we improve the pretraining process and the linear classification of the models in Table~\ref{tab:overall}. For the pretraining process of these old result in this table, please check 
our preprint paper~\url{https://arxiv.org/abs/2108.07582}.}
\label{tab:fewshot}
\end{table}

\subsection{Why geometric and mixup performs better on UAV images?}

\begin{figure}
     \centering
     \begin{subfigure}[b]{0.45\textwidth}
         \centering
         \includegraphics[width=\textwidth]{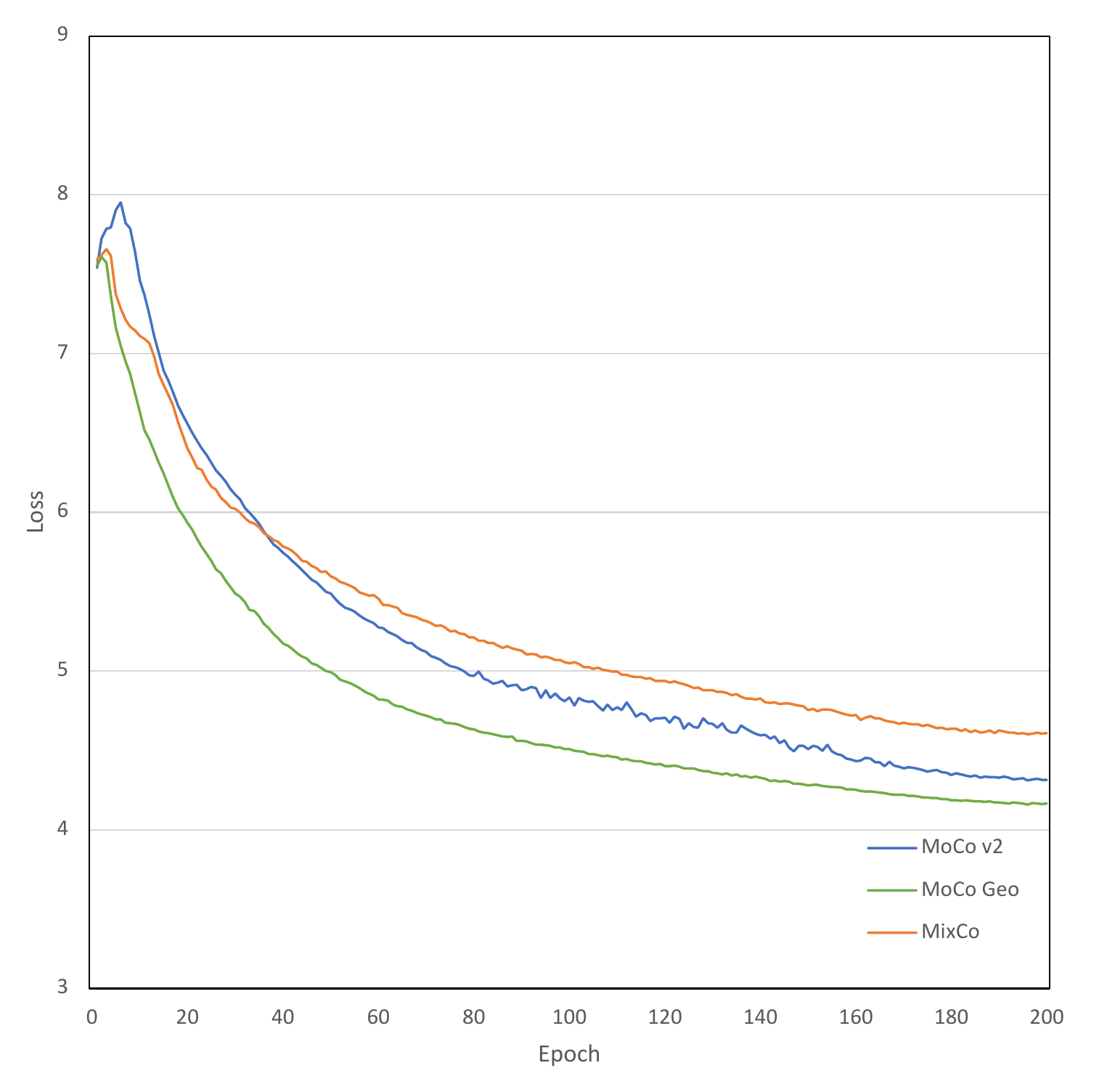}
         \caption{Loss of MoCo v2, MoCo Geo, and MixCo}
         \label{mocoloss}
     \end{subfigure}
     \hfill
     \begin{subfigure}[b]{0.45\textwidth}
         \centering
         \includegraphics[width=\textwidth]{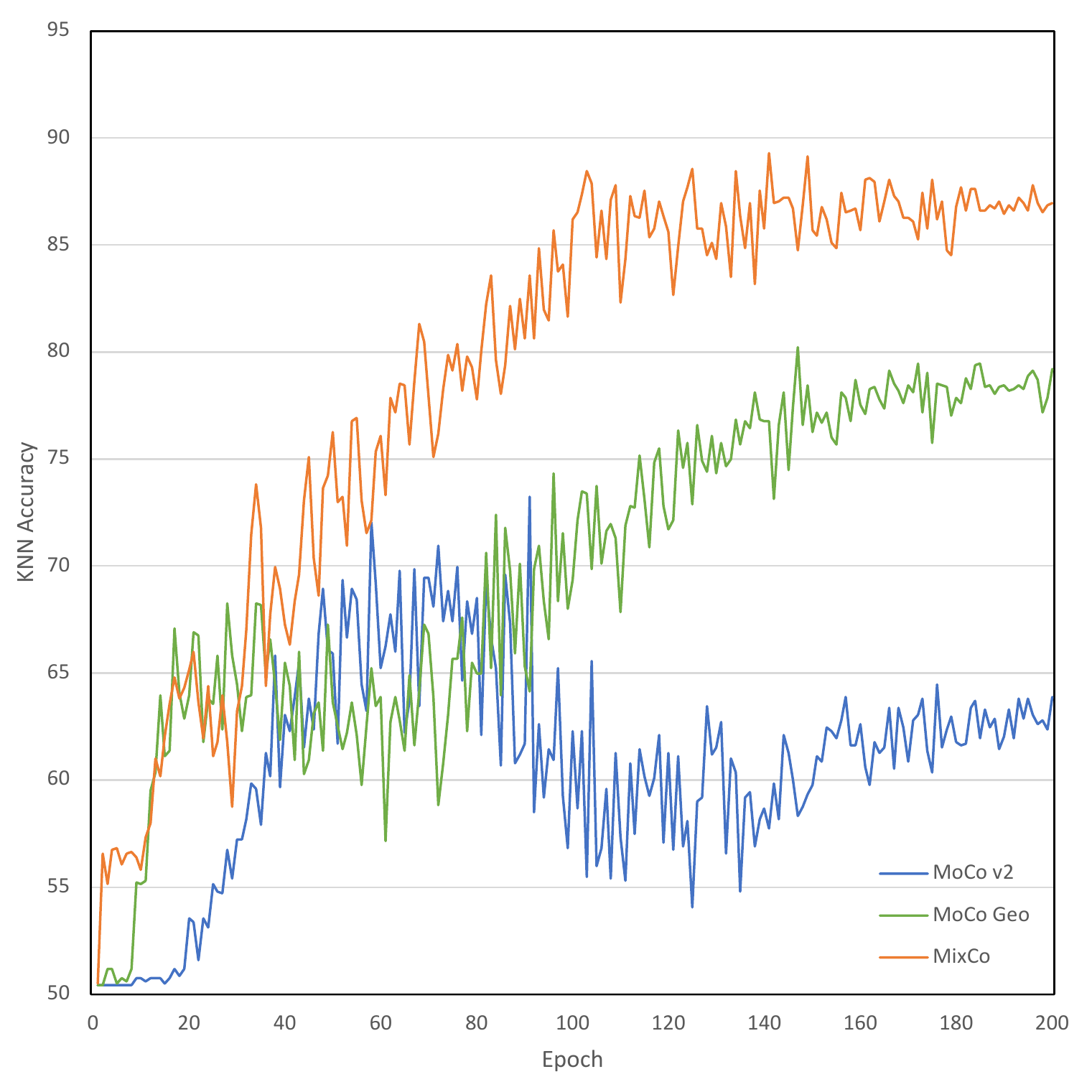}
         \caption{kNN accuracy of MoCo v2, Geo, and MixCo}
         \label{mocoknn}
     \end{subfigure}
     \hfill
     \begin{subfigure}[b]{0.45\textwidth}
         \centering
         \includegraphics[width=\textwidth]{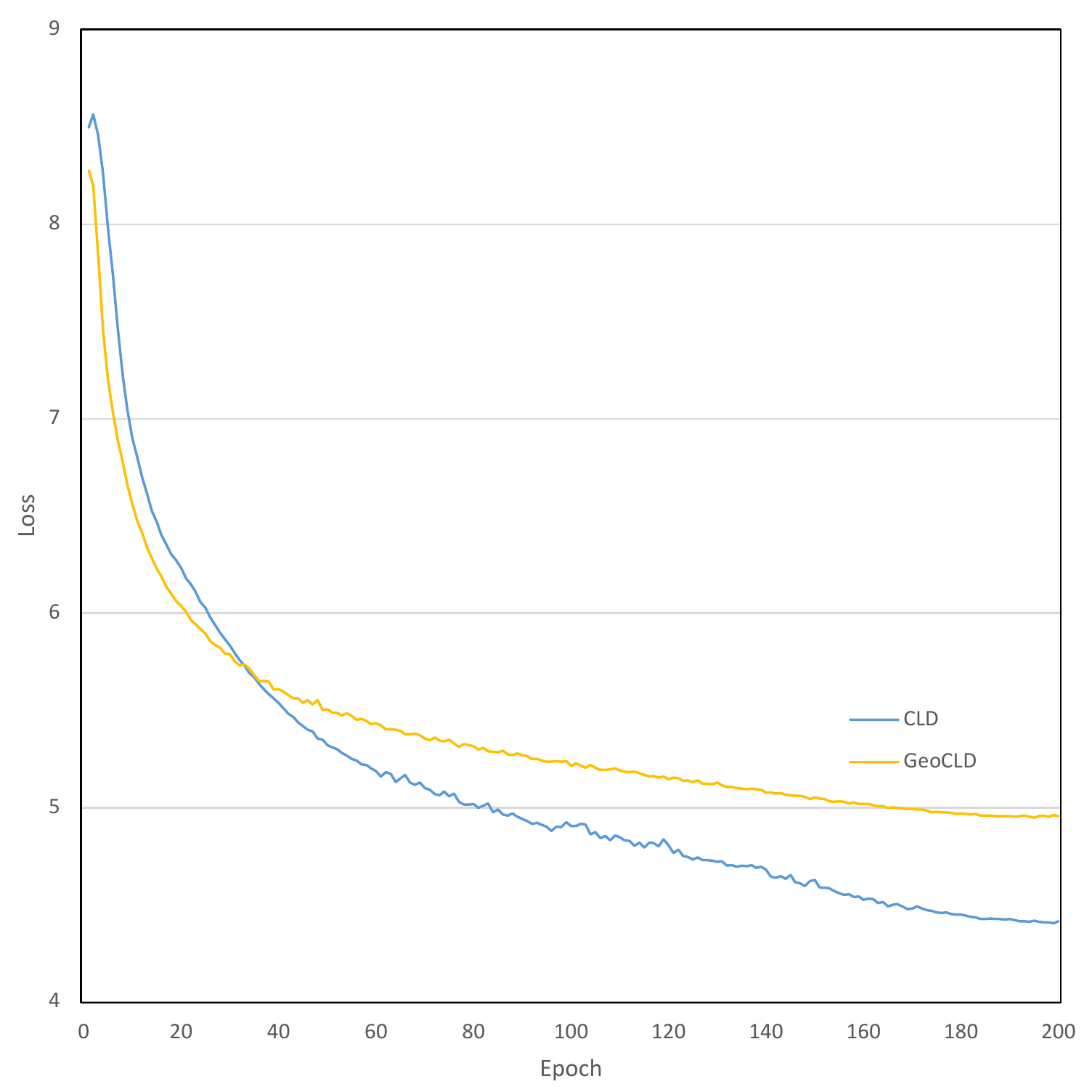}
         \caption{Loss of MoCo+CLD and GeoCLD}
         \label{cldloss}
     \end{subfigure}
     \hfill
     \begin{subfigure}[b]{0.45\textwidth}
         \centering
         \includegraphics[width=\textwidth]{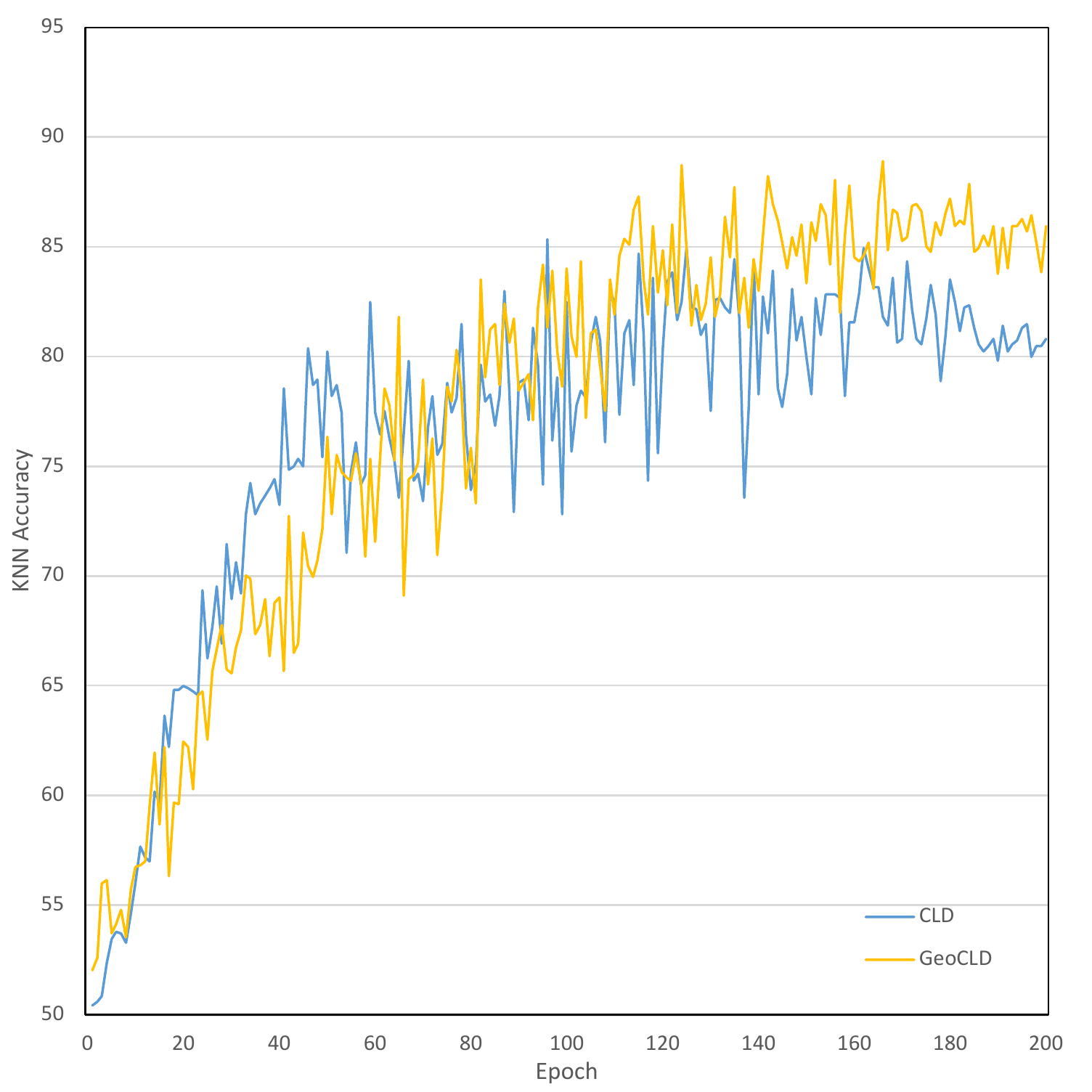}
         \caption{kNN accuracy of MoCo+CLD and GeoCLD}
         \label{cldknn}
     \end{subfigure}
     \hfill
     
\caption{\textbf{Monitoring loss and KNN accuracy} of all models during the pretraining.}
\label{fig:mocomonitor}
\end{figure}

For kNN monitor, we perform a grid search and find the optimal $k$, $t$ are 20 and 0.02 for our KWD-LT dataset. Comparing the kNN accuracies of MoCo v2 with MoCo Geo in Figure~\ref{mocoknn}, MoCo+CLD with GeoCLD in Figure~\ref{cldknn}, we can find that adding geometric augmentation can prevent overfitting.

Mixing the image with its different views will increase the amount of objects in the final image. Thus mixup strategies can reduce the imbalance between positive and negative samples. Rotation is one kind of stronger augmentations. The augmentation methods in MoCo v2~\cite{mocov2} are mainly about texture transformation (color augmentation in Table~\ref{tab:aug}), which will capture content information, \emph{e.g.}, color change. However, our data have very similar pattern and the objects of interest have very similar colors and are very tiny in the raw images. Thus, rotation augmentation will help models "perceive" the orientation change of tiny objects further capture the invariant information.

\subsection{Why CLD performs better on class imbalanced dataset?}
CLD~\cite{Wang_2021_CVPR} groups similar samples and then perform the contrastive mechanism. Clustering instead of directly comparing instances reduces the imbalance between positive and negative samples. For instance, in MoCo, the dictionary size is 4,096, which means each input instance is compared against 4,096 negative samples. After performing local clustering, there are only $k$ clusters, which means each input instance only needs comparing with $k$ negative samples (centroids of $k$ clusters). \textit{The importance of positives increases from 1/4096 to 1/k}~\cite{Wang_2021_CVPR}. CLD thus has better performance regarding invariant mapping for both head and tail classes. 

\section{Longer Training}

We train MoCo v2 and MixCo for 800 epoch. The results show that both models benefit from a longer training compared to its supervised counterpart. In supervised learning, longer training might cause the problem of overfitting. The reason is that training longer provides more negative examples, which can futher improve the results~\cite{SimCLR}. Furthermore, MixCo converges faster than MoCo and longer training might cause overfitting. The results in Table~\ref{tab:longtrain} are slightly different from the results in Table~\ref{tab:overall}. This is caused by same initial learning rate and different training epochs in the learning rate cosine decay  strategy.

\begin{table*}
\begin{center}
\begin{tabular}{llcccccc}
\specialrule{.1em}{.05em}{.05em} 
Model & Epochs & 100 & 200 & 300 & 600 & 800 \\
\hline
MoCo v2 & Linear Acc. & 83.1 & 79.0 & 84.6 & 87.0 & 90.0 \\
MixCo (ours.) & Linear Acc. & 89.6 & 92.0 & 92.9 & 94.0 & 93.8 \\

\specialrule{.1em}{.05em}{.05em} 
\end{tabular}
\end{center}
\caption{Linear accuracy (10\% labels) for long time training.}
\label{tab:longtrain}
\end{table*}

\begin{figure}
     \centering
     \begin{subfigure}[b]{0.48\textwidth}
         \centering
         \includegraphics[width=\textwidth]{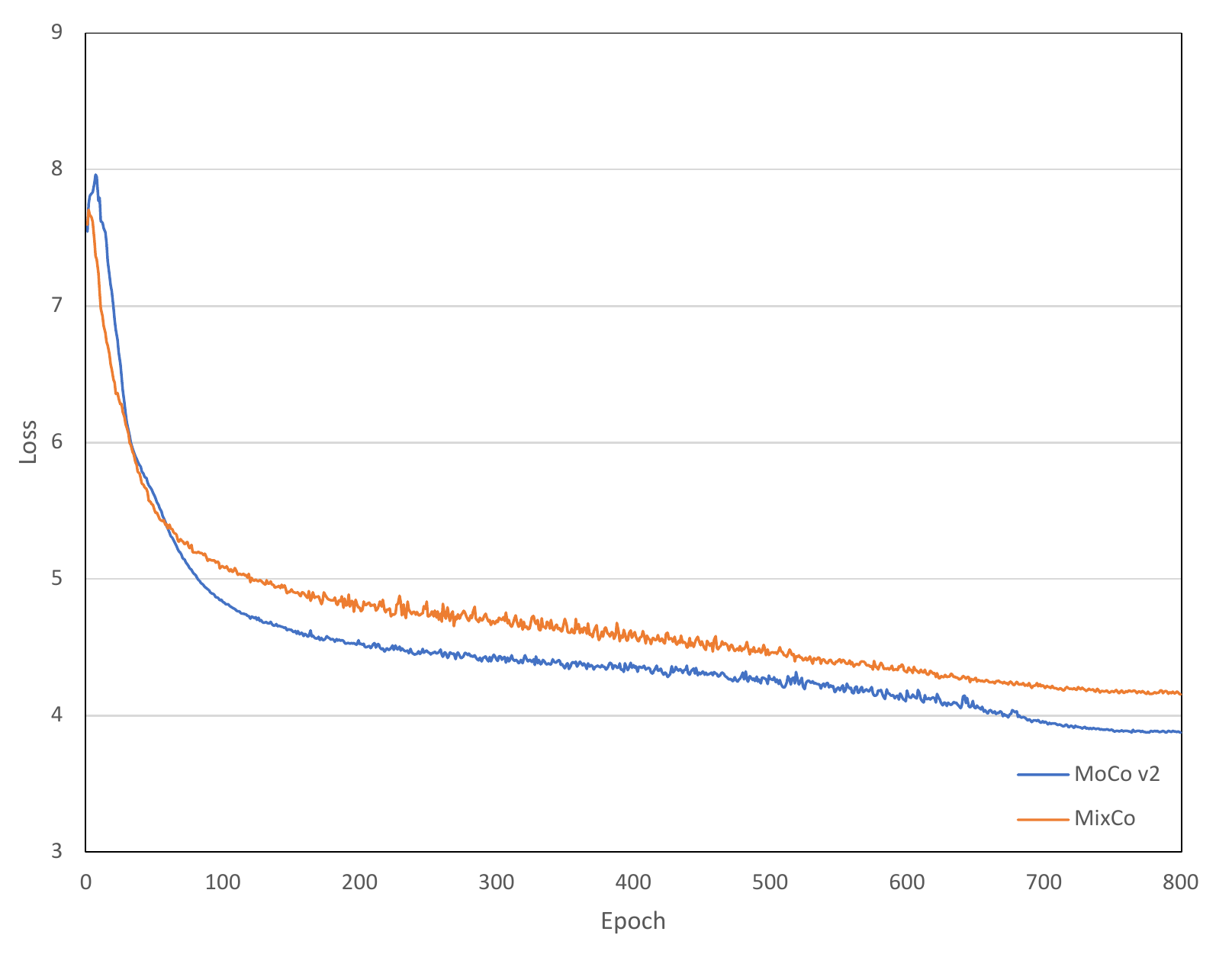}
         \caption{The loss of MoCo v2 and MixCo.}
         \label{loss}
     \end{subfigure}
     \hfill
     \begin{subfigure}[b]{0.46\textwidth}
         \centering
         \includegraphics[width=\textwidth]{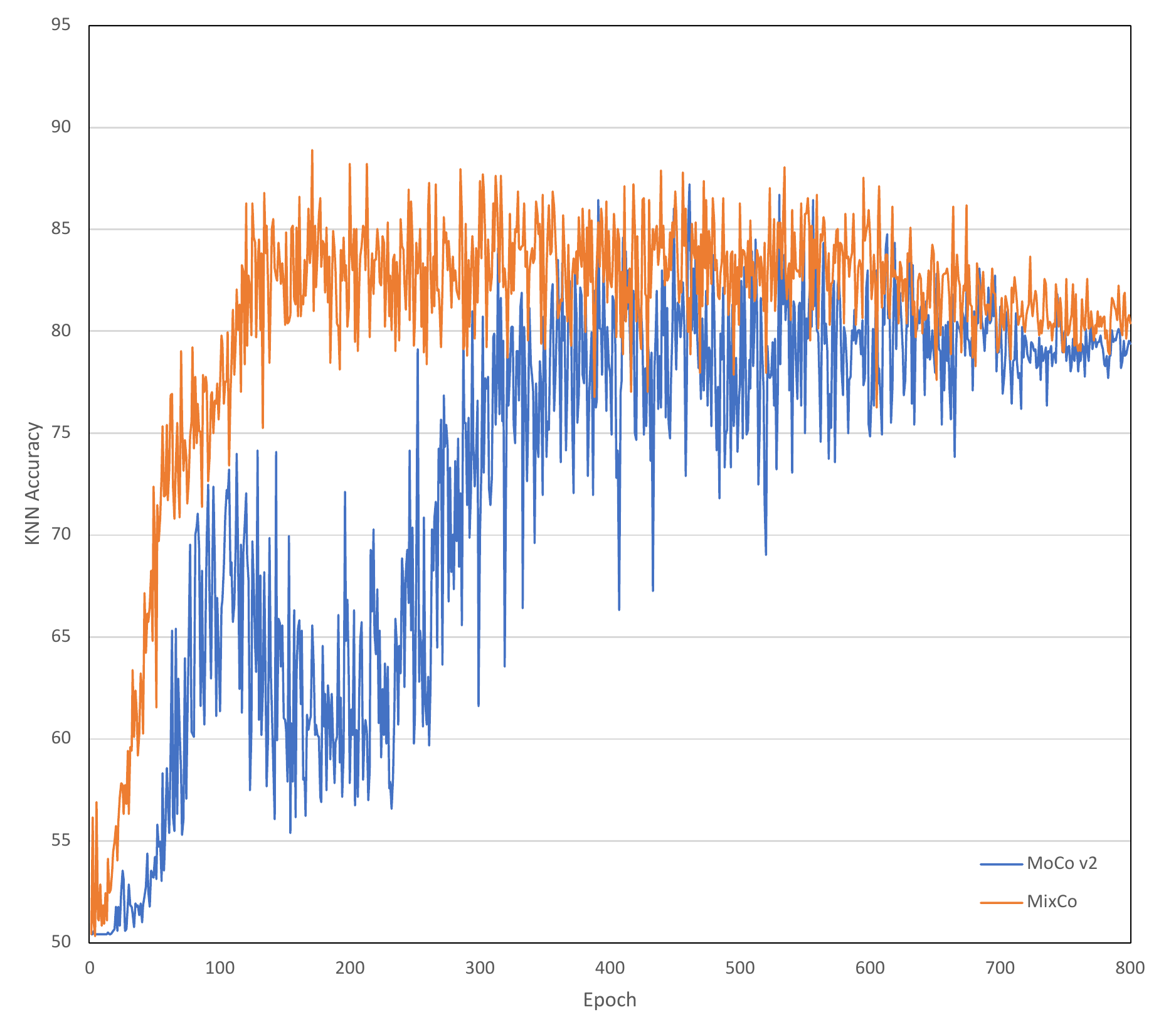}
         \caption{The KNN accuracy of MoCo v2 and MixCo.}
         \label{knnacc}
     \end{subfigure}
\caption{\textbf{Monitoring loss and KNN accuracy} for a longer training.}
\end{figure}

\subsection{Ablation Study}

The results for the ablation studies on the number of clusters for CLD and $\lambda$ are shown in Tables~\ref{tab:lambda} and~\ref{tab:cluster}. We can clearly see that the model performs almost equally when $\lambda = 0.25$ or $\lambda = 1$. Hence, we chose the default $\lambda = 0.25$ value as in~\cite{Wang_2021_CVPR}. However for the number of clusters, there does not seem to be an obvious trend. we speculate that (1) grouping projected \emph{representations} by $k$-means clustering is hard to perform well, (2) the model is not able to recognize animals beneath the tree or/and to distinguish between dead tree trunks and animals (number of clusters = 16 or 32). When number of clusters = 16, 32, or 64, the accuracy is equally good. Therefore, we chose the best recall with number of cluster = 32.

For MixCo, we exploit the effect of different ratio of original MoCo v2 branch and mixup probability. The results have shown that with Geo:Color = 1:9 and 30\% mixup probability, the model has the best performance on KWD dataset with only 10\% labels.

\begin{table}
\begin{center}
\begin{tabular}{lcccc}
\specialrule{.1em}{.05em}{.05em} 
Model & $\lambda$ & Acc & Prec & Rec \\
\hline
MoCo+CLD & 0.1 & 86.2 & 98.6 & 73.4\\
MoCo+CLD & \textbf{0.25} & \textbf{88.4} & 98.9 & \textbf{77.4}\\
MoCo+CLD & 0.5 & 86.4 & 98.9 & 73.5 \\
MoCo+CLD & 1 & 88.1 & 98.5 & 77.2\\
CLD & 1 & 60.9 & 98.9 & 22.1\\
\specialrule{.1em}{.05em}{.05em} 
\end{tabular}
\end{center}
\caption{Hyper-parameter $\lambda$ selection. All models are pretrained on KWD-Pre using the MoCo+CLD strategy.}
\label{tab:lambda}
\end{table}

\begin{table}
\begin{center}
\begin{tabular}{lcccc}
\specialrule{.1em}{.05em}{.05em} 
Model & Clusters & Acc & Prec & Rec \\
\hline
MoCo+CLD & \textbf{16} & 88.3 & 100.0 & 76.5 \\
MoCo+CLD & 30 & 87.3 & 99.7 & 75.0 \\
MoCo+CLD & \textbf{32} & \textbf{88.4} & 98.9 & \textbf{77.4}\\
MoCo+CLD & 48 & 87.2 & 99.7 & 74.9\\
MoCo+CLD & \textbf{64} & \textbf{88.4} & 100.0 & 76.9\\
\specialrule{.1em}{.05em}{.05em} 
\end{tabular}
\end{center}
\caption{Number of clusters selection. All models are pretrained on KWD-Pre using the MoCo+CLD strategy with $\lambda = 0.25$.}
\label{tab:cluster}
\end{table}




\begin{table}[htbp]
    \begin{subtable}[htbp]{0.45\textwidth}
        \centering
        \begin{tabular}{lcccc}
        \specialrule{.1em}{.05em}{.05em} 
        Model & $p$ & Acc & Prec & Rec \\
        \hline
        MixCo & 0 & 92.172 & 98.121 & 85.894 \\
        MixCo & 0.1 & 90.488 & 94.366 & 85.867 \\
        MixCo & 0.3 & 92.761 & 98.859 & 86.448 \\
        MixCo & 0.5 & 91.498 & 96.390 & 86.100 \\
        MixCo & 0.7 & 90.572 & 99.424 & 81.469 \\
        MixCo & 0.9 & 90.320 & 97.000 & 83.045\\
        MixCo & 1 & 91.582 & 98.199 & 84.625 \\
        \specialrule{.1em}{.05em}{.05em} 
        \end{tabular}
        \caption{$\gamma=0.5$, different mixup-p.}
        \label{tab:mixupp}
    \end{subtable}
    \hfill
    \begin{subtable}[htbp]{0.45\textwidth}
        \centering
        \begin{tabular}{lcccc}
        \specialrule{.1em}{.05em}{.05em} 
        Model & $\gamma$ & Acc & Prec & Rec \\
        \hline
        MixCo & 0 & 88.889 & 94.719 & 82.264 \\
        MixCo & 0.3 & 90.152 & 91.166 & 88.718 \\
        MixCo & 0.5 & 92.761 & 98.859 & 86.448 \\
        MixCo & 0.6 & 92.172 & 99.046 & 84.945\\
        MixCo & 0.9 & 94.192 & 99.263 & 89.040 \\
        MixCo & 1 & 91.667 & 99.613 & 83.592 \\
        
        \specialrule{.1em}{.05em}{.05em} 
        \end{tabular}
        \caption{mixup-p=0.3, different $\gamma$.}
        \label{tab:gamma}
     \end{subtable}
     \caption{\textbf{Ablation study} of the MixCo model.}
     \label{tab:ablationmixco}
\end{table}

\section{Conclusion}
In this paper, our proposed strategy reduces the label requirements in the wildlife recognition tasks. The problem is introduced by applying supervised learning to automated animal censuses in largely remote areas with aerial imagery, in which scenario the annotations are expensive to be obtained. The contrastive self-supervised pretraining with domain-specific geometric transformation and mixup strategies outperforms the performance of fine-tuning ImageNet pretrained model with full labels. Results show that the geometric invariant mapping method can capture information more efficiently of wildlife in UAV images than method without geometric augmentation. Extensive experiments further prove the effectiveness of recognizing rare wildlife with reduced labels.

\clearpage

\section{Acknowledgement}

This project would not have been possible without the support of many people. Many thanks to my adviser, Dr. Benjamin Alexander Kellenberger, who read my numerous revisions and helped make some sense of the confusion. His unassuming approach to research and science is a source of inspiration. This approach is reflected by his simple but clear writing style, which is something I hope to carry forward throughout my career. Thanks to my supervisor, Prof. Dr. Devis Tuia. Devis has been an ideal teacher, mentor, and thesis supervisor, offering advice and encouragement with a perfect blend of insight and humor. Also thanks to my supervisor, Prof. Dr. Irena Hajnsek who offered invaluable guidance and support. I’m proud of, and grateful for, my time working with Beni, Devis, and Irena.

Thanks to my dear colleague, Rui Gong. Rui is a good friend and a good teacher. Without his inspired idea, I could not get to my destination. Long live our friendship and wish you all the success with your PhD journey.

Thanks to ETH Zürich for providing me with the financial support to complete this project. And finally, thanks to parents, and numerous friends who endured this long process with me, always offering support and love. I will remember this experience for a life long time.

\clearpage
%
%
\bibliographystyle{splncs04}
\bibliography{xzhengmt}
\end{document}